\documentclass[10pt,journal,cspaper,compsoc]{IEEEtran}
\usepackage[pdftex]{graphicx}
\usepackage{graphicx,epsfig,amsmath}

\usepackage{color}
\usepackage{stmaryrd}
\usepackage{amsmath} 
\usepackage{amssymb} 

\newcommand{\changeBM}[1]{{#1}}

\usepackage[normalem]{ulem}
\newcommand{\Gao}[1]{{#1}}

\ifCLASSOPTIONcompsoc
  \usepackage[nocompress]{cite}
\else
  \usepackage{cite}
\fi
\usepackage{algorithmic}
\ifCLASSOPTIONcompsoc
  \usepackage[caption=false,font=normalsize,labelfont=sf,textfont=sf]{subfig}
\else
  \usepackage[caption=false,font=footnotesize]{subfig}
\fi
\begin{document}
\title{Heterogeneous Tensor Decomposition for Clustering via Manifold Optimization}

\author{Yanfeng~Sun, 
        Junbin~Gao, Xia~Hong, Bamdev~Mishra and
        Baocai~Yin
\IEEEcompsocitemizethanks{\IEEEcompsocthanksitem Yanfeng Sun and Baocai Yin are with Beijing Municipal Key Lab of Multimedia and Intelligent Software Technology, College of Metropolitan Transportation, Beijing University of Technology, Beijing 100124, China. 
E-mail: \{yfsun,ybc\}@bjut.edu.cn
\IEEEcompsocthanksitem Junbin Gao is with School of Computing and Mathematics, Charles Sturt University, Bathurst, NSW 2795, Australia. \protect E-mail: jbgao@csu.edu.au
\IEEEcompsocthanksitem Xia Hong is with the School of Systems Engineering, University of Reading, Reading, RG6 6AY, UK. \protect E-mail:
x.hong@reading.ac.uk
\IEEEcompsocthanksitem Bamdev Mishra is with Department of Electrical Engineering and Computer Science, University of Li\`{e}ge, 4000 Li\`{e}ge, Belgium. \protect E-mail: b.mishra@ulg.ac.be
}
}

\markboth{IEEE Transactions on XXXX,~Vol.~XX, No.~X, April~2015}%
{Sun \MakeLowercase{\textit{et al.}}: Tensor Clustering}
\IEEEcompsoctitleabstractindextext{%
\begin{abstract}
Tensors or multiarray data are generalizations of matrices. Tensor clustering has become a very important research topic due to the intrinsically rich structures in real-world multiarray datasets. Subspace clustering based on vectorizing multiarray data has been extensively researched. \Gao{However, vectorization of tensorial data} \changeBM{does not exploit complete} structure information. In this paper, we propose a subspace clustering algorithm without adopting any vectorization process. Our approach is based on a \changeBM{novel} heterogeneous Tucker decomposition model. \changeBM{In contrast to existing techniques, we propose a new clustering algorithm that alternates between different modes of the proposed heterogeneous tensor model. All but the last mode have closed-form updates. Updating the last mode reduces to optimizing over the so-called \emph{multinomial manifold}, for which we investigate second order Riemannian geometry and propose a trust-region algorithm}. \changeBM{Numerical experiments} show that \changeBM{our proposed algorithm compete effectively with state-of-the-art} clustering algorithms that are based on tensor factorization.


\end{abstract}

}

\maketitle

\IEEEdisplaynotcompsoctitleabstractindextext

\IEEEpeerreviewmaketitle

\section{Introduction}
\IEEEPARstart{I}{n}
the last two decades, the advance of modern sensing, networking, communication and storage technologies {have paved} the way for the availability of multidimensional data with high dimensionality. For example,  remote sensing is producing massive multidimensional data that need to be carefully analyzed. One of the characteristics of these gigantic datasets is that they often have a large amount of redundancies. This motivates the development of  a low-dimensional representation that best assists a range of learning tasks in order to avoid the so-called ``curse of dimensionality'' \cite{ScholkopfSmola2002}. Many data processing tasks involve manipulating \changeBM{multi-dimensional} objects. For example, video data \cite{LuaPlataniotisbVenetsanopoulos2011} can be regarded as an object of pixel location in two dimensions plus one dimension in time. Similar representation can be seen in manipulating remote sensing data \cite{GuoZhangHuang2013}. In analyzing personalized webpages, the data is usually represented as a third order dataset with three dimensional modes of users, query words and webpages\changeBM{,} respectively \cite{SunZengLiuLuChen2005}. In document clustering, one presents the dataset in a three-way format of authors, terms and times \cite{CaiHeHan2006,ForsatiMahdaviShamsfardMeybodi2013}.

The multi-dimensional data are known as tensors \cite{Kolda2006,KoldaBader2009}, where data elements are addressed by more than two indices. An \changeBM{$N$-order} tensor is an element of the tensor product of $N$ vector spaces. A 2D matrix is an example of the 2nd-order tensor. Similarly, hyperspectral imagery \cite{ShawKBurke2003} is naturally a three-dimensional (3D) data cube containing both spatial and spectral dimensions. \changeBM{Simultaneously considering both spectral and spatial structures of hyperspectral data in clustering or classification lead to superior results, e.g., in \cite{ZhangLiDing2013}}. \changeBM{To this end, we prefer treating, e.g., a 3D cube as a whole}. \changeBM{In order to circumvent the issue of high dimensionality, a strategy is to compress the data while capturing the dominant trends or to find the most suitable  ``sparse'' representation of the data.}


Data clustering is one of the widely used data mining techniques \cite{JainMurtyFlynn1999}. Many clustering methods consider each \changeBM{data} as a \changeBM{high dimensional} mathematical vector \changeBM{A typical way is to pre-process those high dimensional vectors with dimensionality reduction techniques such as principal component analysis (PCA) \cite{Bishop2006}.} \changeBM{For dealing with tensorial data, a conventional step} is to \changeBM{first} vectorize them before any analysis is applied. For example, {this procedure is practiced} to convert image data, \changeBM{a} 2nd-order tensor, into vectors for processing. \Gao{Not surprisingly,} this strategy breaks higher order dependencies \changeBM{that may be present} in the data. \changeBM{Recently, new approaches that are capable of directly processing structural information of tensorial data have been proposed, e.g., tensorial data structure is exploited in computer vision applications \cite{Aja-FernandezLuisTaoLi2009,TangSalakhutdinovHinton2013} and machine learning \cite{ShashuaHazan2005,Morup2011}.}



\changeBM{Tensorial} data have a large amount of redundancies. It is desired to have a mechanism to reduce such redundancies for the sake of efficient application of learning algorithms. \changeBM{We use} the general term dimensionality reduction to describe \changeBM{such techniques} or models. There exist various tensor decomposition models, amongst which the CANDECOMP (canonical decomposition)/PARAFAC (parallel factors) or in short \emph{CP decomposition} \cite{Kiers2000} and the \emph{Tucker decomposition} are two fundamental models for tensor decomposition (\changeBM{refer} to the survey paper \cite{KoldaBader2009} for details). \Gao{It should be noted} that the CP decomposition is a special case of \changeBM{the} Tucker decomposition \cite{KoldaBader2009}\changeBM{,} where \changeBM{the} factor matrices have same number of columns and the core tensor is superdiagonal, which means that every mode of the tensor is of the same size and its elements remain constant under any permutation of the indices. Many other decomposition algorithms/models can be viewed as special formats of the CP \changeBM{and} Tucker decomposition. \changeBM{For example, the higher order SVD algorithm (HOSVD) \cite{DeLathauwerDeMoorVandewalle2001}, an extension of the classical SVD,} is a special case of the general Tucker decomposition \changeBM{of a tensor} in which the core tensor is of the same dimension as the tensor to be decomposed and all the mode matrices have \emph{orthonormal columns}. \changeBM{Similarly,} the classical PCA has several extensions \changeBM{for tensorial data}. The generalized tensor PCA (GND-PCA) seeks a shared Tucker decomposition for all the given tensors in which the core tensors are different but the matrix factors along each mode are orthogonal. This decomposition procedure is also called the \emph{higher-order orthogonal iteration} (HOOI) algorithm in \cite{DeLathauwerDeMoorVandewalle2000}.

In applications where data are non-negative such as images, the non-negative matrix factorization (NFM) has proven to be a successful approach for detecting essential features in the data \cite{LeeSeung1999}. Several efficient algorithms have been proposed \cite{LeeSeung2001,Lin2007}. \changeBM{Recently,} NFM has been extended to non-negative tensor factorization (NTF) and \changeBM{has been} investigated in \cite{ShashuaHazan2005,WellingWeber2001,FriedlanderHatz2008,LiuLiuWonkaYe2012,WuTanYangTaoTangZhuang2013}.

Another trend in tensor decomposition research is to introduce more structures in \changeBM{the decomposition model itself}. Recently\changeBM{,} Zhang et al. \cite{ZhangLiDing2013} considered Tri-ONTD (Tri-factor orthogonal non-negative tensor decomposition), \changeBM{a new tensor decomposition model}. The fundamental aim \changeBM{of this model} is to discover common characteristics of a series of matrix data. A straightforward application of Tri-ONTD is to identify cluster structures of \changeBM{a} dataset. The core idea behind this model is based on the centroid-based clustering algorithms such as the \emph{k-means} algorithm. The idea of introducing new structures \changeBM{in tensor decomposition} can also be seen in \cite{LiuWuLiCaiHuang2012} \changeBM{in the context of image representation}.


\changeBM{Often tensor clustering tasks are formulated as optimization problems on specific tensor decomposition models \cite{VichiRocciKiers2007,AcarYener2009,PengLi2011a,SunGaoHongGuoHarris2014}.} \changeBM{For example,} when \changeBM{imposing} some specific constraints, like orthogonality in the HOOI algorithm \cite{DeLathauwerDeMoorVandewalle2000}, the resulting problems \changeBM{reduce to} optimization over \emph{matrix manifolds} \cite{AbsilMahonySepulchre2008}. \changeBM{In the case of the HOOI algorithm for the HOSVD decomposition, the resulting optimization problem is on the \emph{Stiefel manifold} \cite[Section~3.3]{AbsilMahonySepulchre2008})}. \changeBM{While the optimization problem in HOSVD admits a closed-form solution under the least squared error criterion, computing a closed-form solution is not possible for the tensor clustering problem} that we consider in this paper. Most existing algorithms avoid this issue by \changeBM{reformulating} it into an optimization problem over the ``flat'' Euclidean space with some treatment, e.g., \changeBM{by introducing a regularization term}.

Recent years have witnessed \changeBM{significant development of} Riemannian optimization algorithms on matrix manifolds such as the Stiefel manifold, the Grassmann manifold, and the manifold of positive definite matrices \cite{AbsilMahonySepulchre2008, Vandereycken2010, Meyer2011}. \changeBM{The Riemannian optimization framework endows a matrix manifold constraint with a \emph{Riemannian manifold structure}. Conceptually, it translates a constrained optimization problem into an \emph{unconstrained optimization} problem on a Riemannian manifold.} Since the Riemannian optimization \changeBM{framework} is directly based on nonlinear manifolds, one can eliminate those constraints such as orthogonality to obtain an unconstrained optimization problem that, by construction, will only use feasible points. The recent successful applications of the Riemannian optimization \changeBM{framework} in machine learning, computer vision and data mining, include low rank optimization \cite{MishraMeyerBachSepulchre2013, Vandereycken2010, Meyer2011}, estimation \cite{Boumal2014}, Riemannian dictionary learning \cite{XieHoVemuri2013,HarandiHartleyShenLovellSanderson2014}, and computer vision tasks \cite{Lui2012}, \changeBM{to name a few}.

In this paper, \changeBM{we propose a novel subspace clustering algorithm that exploits the tensorial structure of data. To this end, we introduce a new heterogeneous Tucker decomposition model.} \changeBM{The proposed clustering algorithm alternates between different modes of the proposed heterogeneous tensor model. All but the last mode have closed-form updates. Updating the last mode reduces to optimizing over the \emph{multinomial manifold}, defined in Section \ref{Manifold}, for which we investigate second order Riemannian geometry and propose a trust-region algorithm. The multinomial manifold is given a Riemannian manifold structure by endowing it with the \emph{Fisher information metric} \cite{InokuchiMiyamoto2007, LebanonLafferty2004}. The Fisher information metric gives the multimonial manifold a differentiable structure, i.e., it ensures that the boundary is ``\Gao{scaled}'' to infinity.}




\changeBM{The} contribution of this paper is \Gao{twofold}. First, we propose a heterogeneous Tucker decomposition model for tensor clustering. Second, we investigate the Riemannian geometry of the multinomial manifold and apply the Riemannian trust-region algorithm to the resulting nonlinear clustering problem over the multinomial manifold.

\changeBM{The paper is organized as follows. We introduce the notations for tensor representation and operations used in this paper in Section \ref{Sec:2.1}. Section \ref{Sec:2.2} introduces the clustering scheme based on the proposed heterogeneous Tucker decomposition model. The associated optimization problems are discussed in Section \ref{Sec:2.3}. Section \ref{Manifold} explores the Riemannian geometry for the multinomial manifold and develops all the necessary optimization-related ingredients. Section \ref{Sec:4} presents the algorithm procedure for the tensor clustering including the proposed Riemannian trust-region algorithm. Section \ref{Sec:5} shows numerical experimental results on both synthetic tensorial data and real-world datasets. Finally Section \ref{Sec:6} concludes the paper.}

\section{Heterogeneous Tucker decomposition model for clustering}

\changeBM{In this section, starting with notation for tensors, we motivate the work, and finally propose the new heterogeneous Tucker decomposition model for clustering.}

\subsection{Tensor Notation and Operations}\label{Sec:2.1}
In the sequel, we follow the \changeBM{convention} used in \cite{KoldaBader2009} to denote 1D vector by lowercase \Gao{boldface} symbols like $\mathbf v$, 2D matrix by uppercase \Gao{boldface} symbols like $\mathbf U$ and general tensors by calligraphy symbols like $\mathcal{X}$.

Let $\mathcal{X}\in\mathbb{R}^{I_1\times \cdots \times I_n\times \cdots \times I_{N}}$ be an $N$-order tensor with $x_{i_1\cdots i_n\cdots i_{N}}$ as the $(i_1\cdots i_n\cdots i_{N})$th element.  The $n$-mode product of an $N$-order tensor $\mathcal{X}$ with a matrix \changeBM{$\mathbf U_n\in\mathbb{R}^{I_n\times J_n}$} is denoted by $\mathcal{X}\times_n \mathbf U_n$. The result is an $N$-order tensor of dimension $I_1\times \cdots \times I_{n-1}\times J_n\times I_{n+1}\times \cdots \times I_{N}$. Element wise, the $n$-mode product \changeBM{is} expressed as
\[
(\mathcal{X}\times_n \mathbf U_n)_{i_1\cdots i_{n-1}j_ni_{n+1}\cdots i_{N}} = \sum^{I_n}_{i_n=1}x_{i_1\cdots i_{n-1}i_ni_{n+1}\cdots i_{N}} u_{j_ni_n}.
\]

\changeBM{Given} an $N$-order tensor $\mathcal{X}\in\mathbb{R}^{I_1\times \cdots \times I_n\times \cdots \times I_{N}}$, we seek a Tucker model, as defined below, to approximate the tensor \changeBM{$\mathcal{X}$ as}
\begin{align}
\mathcal{X} \approx &\, \mathcal{G}\times_1 \mathbf U_1\times_2 \mathbf  U_2 \times_3\cdots \times_N \mathbf  U_N \notag \\
\triangleq &\, \llbracket \mathcal{G}; \mathbf U_1, \mathbf U_2, ..., \mathbf U_N\rrbracket,  \label{TuckerDef}
\end{align}
where $\mathcal{G}$ is an $N$-order tensor of dimension $J_1\times \cdots \times J_n\times \cdots \times J_{N}$ with $J_n\leq I_n$, called the core tensor, and \changeBM{$\mathbf U_n\in\mathbb{R}^{I_n\times J_n}$} is the matrix applied along mode-$n$. In this decomposition, \changeBM{the core tensor $\mathcal{G}$ is interpreted as} a lower dimensional representation of the tensor $\mathcal{X}$. The Tucker decomposition is a form of higher-order PCA \cite{KapteynNeudeckerWansbeek1986}, where all the factor matrices $\mathbf U_n$ are shared by a group of given tensors.

\subsection{Heterogeneous Tucker Decomposition Model}\label{Sec:2.2}
Most Tucker decomposition models are of a homogeneous nature, by which we mean that all the factor matrices $\mathbf{U}_n$ \changeBM{satisfy the same constraint}. For example, in the \changeBM{classical} HOSVD \cite{KoldaBader2009}, all the factor matrices are required to be orthogonal, i.e., $\mathbf U^T_n\mathbf U_n=\mathbf I_{J_n} $ (denoted by $\mathbf I$ for simplicity). In the nonnegative Tucker decomposition \changeBM{model} \cite{FriedlanderHatz2008,LiuLiuWonkaYe2012}, all the factors are \changeBM{matrices} with nonnegative entries, i.e., $\mathbf U_n \geq 0$.

However, the \changeBM{requirement for homogeneous factors  $\mathbf{U}_n$ is not preferred in many cases}. \changeBM{Especially, when} the factor matrices in different modes have different interpretations. \Gao{For example, consider the problem of clustering a set of images (2-order tensors). We can stack all the images onto a 3-order tensor, \changeBM{where the third mode corresponds to the number of images}. For clustering \changeBM{the images}, \changeBM{it is of interest} to decompose the entire 3-order tensor in a way that, along the third mode, each image \changeBM{is} represented by several cluster representatives, as done in fuzzy k-means algorithms. This can be achieved by \changeBM{ensuring that} the last mode factor matrix, \changeBM{i.e.,} $\mathbf U_3$ \changeBM{is} \changeBM{nonnegative} and the row sum of $\mathbf U_3$ \changeBM{is} $1$ to mimic the cluster probability of an image.}

In general, we suppose that we are given a set of $M$ $(N-1)$-order tensors, denoted by $\{\mathcal{X}_1, \mathcal{X}_2, ..., \mathcal{X}_{M}\}$ and we want cluster them into $K$ clusters. \changeBM{This can be done by projecting the tensors along all the first $N-1$ modes}, then cluster them. \changeBM{To this end,} we stack all the $M$ $(N-1)$-order tensors along the $N$ mode, so that we \changeBM{have} an $N$-order tensor $\mathcal{X}$ \changeBM{in a way} that each slice of $\mathcal{X}$, along the last mode, is one of $(N-1)$-order \changeBM{tensors} $\mathcal{X}_l$ ($l=1,2,...,M$). \changeBM{Following the general notation in Section \ref{Sec:2.1}, we have $I_N = M$}.

Our proposed model is defined by the \changeBM{optimization problem}
\begin{align}
\min_{\mathcal{G}, \mathbf U^T_1\mathbf U_1=\mathbf I, ..., \mathbf U^T_{N-1}\mathbf U_{N-1} = \mathbf I, \mathbf U_N\mathbf{1} = \mathbf{1}, \mathbf U_N\geq 0}f(\mathcal{G}, \mathbf U)
\label{30March2014-1}
\end{align}
with
\[
f(\mathcal{G}, \mathbf U) = \frac12\|\mathcal{X} - \mathcal{G}\times_1 \mathbf U_1\times_2 \mathbf U_2\times_3 \cdots \times_N \mathbf U_N\|^2_F,
\]
where $\mathbf 1$ is a column vector of all ones, $\mathbf U_1, ..., \mathbf U_{N-1}$ are matrices whose columns are orthogonal, \changeBM{$\| \cdot\|_F$ denotes the \emph{Frobenius} norm, and} $\mathbf U_N \in \mathbb{R}^{M\times K}$ is nonnegative and the sum of each row is $1$. \changeBM{In (\ref{30March2014-1})}, the dimension of $\mathcal{G}$ in mode $N$ is $K$.

\changeBM{If each} of the $K$ slices of $\mathcal{G}$ \changeBM{is interpreted} as \emph{cluster centroids} in the projected space $\mathbb{R}^{J_1\times J_2\times\cdots \times J_{N-1}}$, then the rows of $\mathbf U_N$ \changeBM{has the interpretation of} cluster indicators.

\changeBM{Given this heterogeneous Tucker decomposition model that is specifically aimed  for clustering, we have a new type of matrix manifold, that is the cartesian product of the Stiefel manifolds, corresponding to the constraints $\mathbf U^T_1\mathbf U_1=\mathbf I, ..., \mathbf U^T_{N-1}\mathbf U_{N-1} =\mathbf I$ and the the \emph{multinomial manifold} that corresponds to $\mathbf U_N\mathbf{1} = \mathbf{1}, \mathbf U_N\geq 0$.} \Gao{It should be noted} that the problem \changeBM{(\ref{30March2014-1})} needs two different treatments in terms of optimizing their factor matrices, i.e., for the first   $(N-1)$ orthogonal matrices  and the tensor clustering indicator matrix $\mathbf U_N$, respectively. In the following we formulate these optimization problems.

\subsection{Optimization}\label{Sec:2.3}

To reduce the number of optimization variables in problem \eqref{30March2014-1}, let us optimize the core tensor $\mathcal{G}$ when all the factor matrices are fixed to their current values. This becomes a \changeBM{least-squares} problem. It is \Gao{straightforward} to prove that the solution is given by \cite{KoldaBader2009}
\begin{align}
\mathcal{G} = \mathcal{X}\times_1 \mathbf U^T_1 \times_2 \cdots \times_{N-1} \mathbf U^T_{N-1}\times_N [(\mathbf U^T_N\mathbf U_N)^{-1} \mathbf U^T_N],  \label{30March2014-2}
\end{align}
where we have used the orthogonality of $\mathbf U_n$ ($n=1,2,...,N-1$). \Gao{It should be noted} that $\mathbf U_N$ \changeBM{is} column full rank.

\changeBM{Using} \eqref{30March2014-2} in the objective function \changeBM{of} \eqref{30March2014-1} results in the following relation
\begin{align*}
 f& = \frac12\|\mathcal{X} - \mathcal{G}\times_1 \mathbf U_1\times_2 \mathbf U_2\times_3 \cdots \times_N \mathbf U_N\|^2_F\\
  & =  \frac12\|\mathcal{X}\|^2_F - \frac12\langle \mathcal{G}\times_N (\mathbf U^T_N \mathbf U_N), \mathcal{G}\rangle\\
  & = \frac12\|\mathcal{X}\|^2_F - \frac12\langle \mathcal{G}\times_N  \mathbf U_N, \mathcal{G} \times_N \mathbf U_N\rangle \\
  & = \frac12\|\mathcal{X}\|^2_F - \frac12 \|\mathcal{X}\times_1 \mathbf U^T_1 \times_2 \cdots \times_{N-1} \mathbf U^T_{N-1}\times_N\\
 & \phantom{ =\frac12\|\mathcal{X}\|^2_F - \frac12 \|\mathcal{X}\times_1}\times_N [\mathbf U_N(\mathbf U^T_N\mathbf U_N)^{-1} \mathbf U^T_N]\|^2_F.
\end{align*}

\changeBM{Subsequently,} minimizing \eqref{30March2014-1} \changeBM{is equivalent to}
\begin{align}
\max_{\mathbf U^T_1\mathbf U_1=\mathbf I, ..., \mathbf U^T_{N-1}\mathbf U_{N-1} = \mathbf I, \mathbf U_N\mathbf{1} = \mathbf{1}, \mathbf U_N\geq 0} \quad h(\mathbf U),
\label{30March2014-4}
\end{align}
where
\begin{align*}
 h(\mathbf U)= \frac12\|&\mathcal{X}\times_1 \mathbf U^T_1 \times_2 \cdots  \\
  &\times_{N-1} \mathbf U^T_{N-1}\times_N [\mathbf U_N(\mathbf U^T_N\mathbf U_N)^{-1} \mathbf U^T_N]\|^2_F.
\end{align*}

\changeBM{It should be emphasized that the function $h$ is \emph{smooth} in the variables and the constraints are \emph{separable}. This motivates to consider an \emph{alternating optimization scheme}} for \eqref{30March2014-4}, \changeBM{where we} maximize \eqref{30March2014-4} with respect to one variable while fixing others. This procedure is cyclically repeated.

Let $\mathbf V_N = \mathbf U_N(\mathbf U^T_N\mathbf U_N)^{-1} \mathbf U^T_N$. \changeBM{If all but $\mathbf U_n$ ($n=1,2,...,N-1$) are fixed, then $\mathbf U_n$ is optimized by solving an \emph{eigenvector problem}.} Specifically, consider updating $\mathbf U_n$ ($n=1, 2, ..., N-1$) while all the others being fixed. Denote
\[
{\mathbf U}_{(-n)} = \mathbf V_{N}\otimes \cdots \otimes \mathbf U_{(n+1)}\otimes \mathbf U_{(n-1)}\otimes \cdots \otimes \mathbf U_{1}.
\]
Subsequently, \eqref{30March2014-4} can be rewritten as, after mode-$n$ matricization,
\begin{align}
\max_{\mathbf U^T_n\mathbf U_n=I_n} \quad \frac12\|\mathbf U_n (\mathbf X_{(n)}\mathbf U^T_{(-n)})\|^2_F, \label{31March2014-5}
\end{align}
where $\mathbf X_{(n)}$ is the $n$-mode matricization of tensor $\mathcal{X}$. The problem (\ref{31March2014-5}) has a \changeBM{closed-form} solution given by, letting $\mathbf B_n =\mathbf X_{(n)}\mathbf U^T_{(-n)}$,
\begin{align}
\mathbf U_n = \text{uf}(\mathbf B^T_n),  \label{31March2014-6}
\end{align}
where $\text{uf}(\cdot)$ extracts the orthogonal factor of the polar decomposition of its argument and is computed as \changeBM{$\text{uf}(\mathbf A) = \mathbf P\mathbf Q^T$, where $\mathbf A = \mathbf P\Sigma \mathbf Q^T$ is the thin singular value decomposition of $\mathbf A$}.

As discussed in Section \ref{Sec:2.2}, the \changeBM{nonnegativity constraint on} $\mathbf U_N$ is specifically introduced for clustering, \Gao{leading to the optimization problem}
\begin{align}
\max_{\mathbf U_N\mathbf{1}  = \mathbf{1}, \mathbf U_N\geq 0} & \quad \frac12\|\mathbf V_N \mathbf B_N\|^2_F  = \frac12\textrm{tr}(\mathbf B^T_N \mathbf V^T_N\mathbf V_N \mathbf B_N)\notag\\
 &=\frac12\textrm{tr}(\mathbf B^T_N \mathbf U_N(\mathbf U^T_N\mathbf U_N)^{-1}\mathbf U^T_N \mathbf B_N),
\label{30March2014-5}
\end{align}
where
\[
\mathbf B_N = \mathbf{X}_{(N)}[\mathbf U^T_{N-1}\otimes \cdots \otimes \mathbf U^T_1]^T = \mathbf{X}_{(N)}[\mathbf U_{N-1}\otimes \cdots \otimes \mathbf U_1]
\]
with $\mathbf{X}_{(N)}$ \Gao{as} the $N$-mode matricization of $\mathcal{X}$ and $\otimes$ \Gao{is} the Kronecker product of matrices.

Problem \eqref{30March2014-5} is a nonlinear optimization \changeBM{problem over the constraint} $\{\mathbf U_N | \mathbf U_N\mathbf{1}  = \mathbf{1}, \mathbf U_N\geq 0\}$, called the multinomial manifold, \changeBM{where $\bf 1$ is the vector of all ones}. An efficient numerical algorithm is needed. To this end, we propose an a trust-region algorithm in Section \ref{Sec:4.1}  based on the Riemannian structure of the multinomial manifold that is discussed in Section \ref{Manifold}.

\Gao{It should be noted that there exist several efficient algorithms for linearly constrained smooth \changeBM{convex} optimization, e.g., \cite{TsengYun2015}.} \changeBM{However, it is not clear whether such approaches are efficiently implementable for a structured nonconvex objective function such as the one in (\ref{30March2014-5}).}

\section{The multinomial manifold}\label{Manifold}
For the concepts of general abstract manifolds, we refer readers to the textbook \cite{Lee2002}. Each row of $\mathbf U_N$ is a discrete probability distribution which describes the membership over the tensor centroids. All the discrete probability distributions make up the multinomial manifold (also called a simplex) defined by
\[
\mathbb{P}^K = \left\{\mathbf u=(u_1, ..., u_K)^T\in\mathbb{R}^{K}: \; u_k > 0, \sum^{K}_{k=1}u_k = 1\right\}.
\]
\changeBM{It should be noted that the nonnegativity constraint $ u_k \geq 0$ is replaced with strict positivity to ensure that the set $\mathbb{P}^K $ is differentiable.} \changeBM{A possible use of the multinomial manifold is in proposing a classifier with kernels, e.g., in \cite{LebanonLafferty2004, ZhangChenLee2005, InokuchiMiyamoto2007}.} For the purpose of solving the optimization problem \eqref{30March2014-5}, we investigate the product manifold of multiple multinomial manifolds, still called the multinomial manifold, defined as
\[
\mathbb{P}^K_M = \left\{\mathbf U = [U_{mk}]\in\mathbb{R}^{M\times K}: \; U_{mk} > 0, \sum^{K}_{k=1}U_{mk} = 1\right\}.
\]

\changeBM{Despite the use of the Fisher information metric on the multinomial manifold in \cite{LebanonLafferty2004,InokuchiMiyamoto2007}, to the best of our knowledge, the derivation of the Riemannian gradient and the Riemannian Hessian of a smooth objective function is new. The computations are shown in the subsequent sections}.

\subsection{The Submanifold Structure}\label{sec:submanifold}
Let us represent an element of $\mathbb{P}^K_M$ with the notation $\mathbf U$ which is the matrix representation of size ${M \times K}$. It is straightforward to show that the tangent space at $\mathbf U \in \mathbb{P}^K_M$ is given by
\changeBM{\[
T_{\mathbf U}\mathbb{P}^K_M = \{\eta_{\mathbf U} \in\mathbb{R}^{M\times K}: \eta_{\mathbf U} \mathbf 1 = \mathbf 0\},
\]}where  $\mathbf 1\in\mathbb{R}^K$ is a column vector of all ones and $\mathbf 0\in\mathbb{R}^M$ is a column vector of all zeros.

\Gao{It should also be noted} that, we can characterize the manifold $\mathbb{P}^K_M$ as an \emph{embedded Riemannian submanifold} of the Euclidean space $\mathbb{R}^{M\times K}$ equipped with the metric $g$ (inner product), i.e.,
\begin{align}
g_{\mathbf U}(\xi_{\mathbf U}, \eta_{\mathbf U}) = \sum_{m,k}\frac{(\xi_{\mathbf U})_{mk}(\eta_{\mathbf U})_{mk}}{U_{mk}}, \label{1April2014-1}
\end{align}
where $\xi_{\mathbf U}$ and $\eta_{\mathbf U}$ belong to the tangent space $T_{\mathbf U}\mathbb{P}^K_M$ at the point $\mathbf U$ on the manifold $\mathbb{P}^K_M$. The metric $g_{\mathbf U}$ defining the new Riemannian structure of the manifold is the called \emph{Fisher information metric} \cite{InokuchiMiyamoto2007}. The inner product defined in \eqref{1April2014-1} determines the geometry such as distance, angle, curvature on $\mathbb{P}^K_M$.

The notion of \emph{Riemannian immersion} helps in computing the Riemannian gradient and Hessian formulas on the manifold \changeBM{$\mathbb{P}^K_M$} in terms of their formulas in the embedding space \changeBM{$\mathbb{R}^{M\times K}$}. The basis of this is the following orthogonal, in the sense of the proposed metric, projection operator.

To project a matrix $\mathbf Z\in\mathbb{R}^{M\times K}$ onto the tangent space $T_{\mathbf U}\mathbb{P}^K_M$, we \changeBM{define} the linear operation $\Pi_{\mathbf U}: \mathbb{R}^{M\times K}\changeBM{\rightarrow} T_{\mathbf U}\mathbb{P}^K_M: \mathbf Z \changeBM{\mapsto} \Pi_{\mathbf U}(\mathbf Z)$ as
\[
\Pi_{\mathbf U}(\mathbf Z) = \mathbf Z - (\alpha \mathbf 1^T)\odot \mathbf U,
\]
where $\alpha = \mathbf Z\mathbf 1\in\mathbb{R}^M$ and $\odot$ is \Gao{the element-wise matrix multiplication operation}. This \changeBM{projection operation is} computed by characterizing the tangent space and its complementary space in the sense of the metric (\ref{1April2014-1}).

Another important concept in the recent retraction-based framework of Riemannian optimization is \Gao{the concept of a \emph{retraction} operation} \cite[Section~4.1]{AbsilMahonySepulchre2008}. \changeBM{The retraction mapping is used to locate the next iterate on the manifold along a specified tangent vector \cite[Chapter~4]{AbsilMahonySepulchre2008}.} \changeBM{The \emph{exponential map} $\text{Exp}_{\mathbf U}$ is the canonical choice for the retraction mapping, which generalizes the notion of ``following a straight line'' in the Euclidean space. However, in this paper we work with the following standard approximation that is easier to characterize. Given a tangent vector $\xi_{\mathbf U} \in T_{\mathbf U} \mathbb{P}^K_M$, the proposed retracting mapping is
\begin{align*}
\mathbf U_+ = & R_{\mathbf U}(\xi_{\mathbf U})  \\
:= & (\mathbf U \odot \exp (t (\xi_{\mathbf U} \oslash \mathbf U))) \oslash (\mathbf U \odot \exp ( (\xi_{\mathbf U} \oslash \mathbf U)) \mathbf 1\mathbf 1^T),
\end{align*}
where $\oslash$ is the element-wise matrix inversion and $\exp(\cdot)$ is \Gao{the element-wise exponential operator applied to matrices.}}

\subsection{The Riemannian Gradient Computation}
Let $\text{Grad}F(\mathbf U)$ be the Euclidean gradient of a \changeBM{smooth} function $F:\mathbb{P}^K_M\longmapsto\mathbb{R}$ with the Euclidean metric. The gradient in $\mathbb{R}^{M\times K}$ endowed with the metric $g$ is scaled as $\text{Grad}F(\mathbf U)\odot \mathbf U$. This can be attributed to the ``change of basis'' in the Euclidean space $\mathbb{R}^{M\times K}$.

The expression of the Riemannian gradient $\text{grad}F(\mathbf U)$ \changeBM{on $\mathbb{P}^K_M$} is obtained by projecting the scaled-gradient $\text{Grad}F(\mathbf U)\odot \mathbf U$ to the tangent space $T_{\mathbf U}\mathbb{P}^K_M$, i.e.,
\begin{align}
\text{grad}F(\mathbf U) = \Pi_{\mathbf U}(\text{Grad}F(\mathbf U)\odot \mathbf U).
\label{31March2014-1}
\end{align}

\subsection{The Riemannian Hessian Computation}
\changeBM{In order to compute the Riemannian Hessian, we need the notion of the \emph{Riemannian connection} \cite[Section~5.5]{AbsilMahonySepulchre2008}. The Riemannian connection, denoted as  $\nabla_{\xi_{\mathbf U}}\eta_{\mathbf U}$, generalizes the \emph{covariant-derivative} of the tangent vector $\eta_{\mathbf U}$ along the direction of the tangent vector $\xi_{\mathbf U}$ on the manifold $\mathbb{P}^K_M$.} Since $\mathbb{P}^K_M$ is a Riemannian submanifold of $\mathbb{R}^{M\times K}$, \changeBM{the Riemannian connection on $\mathbb{P}^K_M$} is also characterized by the projection of the corresponding connection $\overline{\nabla}_{\xi_{\mathbf U}}\eta_{\mathbf U}$ in the embedding space $\mathbb{R}^{M\times K}$ endowed with the metric (\ref{1April2014-1}), i.e., $\nabla_{\xi_{\mathbf U}}\eta_{\mathbf U} = \Pi_{\mathbf U}(\overline{\nabla}_{\xi_{\mathbf U}}\eta_{\mathbf U})$ \changeBM{\cite[Proposition~5.3.2]{AbsilMahonySepulchre2008}}.

The connection $\overline{\nabla}$ in the Euclidean space $\mathbb{R}^{M\times K}$ endowed with the metric (\ref{1April2014-1}) is computed using the \emph{Koszul formula} \cite[Theorem 5.3.1]{AbsilMahonySepulchre2008}, and after a few steps of computations, it admits the matrix characterization
\[
\overline{\nabla}_{\xi_{\mathbf U}}\eta_{\mathbf U} = D\eta_{\mathbf U}[\xi_{\mathbf U}] -\frac12(\xi_{\mathbf U}\odot\eta_{\mathbf U})\oslash\mathbf U.
\]

The Riemannian Hessian $\text{Hess}F(\mathbf U)[\xi_{\mathbf U}]$ is the covariant-derivative of the Riemannian gradient $\text{grad}F(\mathbf U)$ in the direction $\xi_{\mathbf U} \in T_{\mathbf U}\mathbb{P}^K_M$, i.e.,
\begin{align}
&\text{Hess}F(\mathbf U)[\xi_{\mathbf U}] =  \Pi_{\mathbf U}(\overline{\nabla}_{\xi_{\mathbf U}}\text{grad}F(\mathbf U))\notag\\
=& \Pi_{\mathbf U}(D\text{grad}F(\mathbf U)[\xi_{\mathbf U}] - \frac12(\xi_{\mathbf U}\odot\text{grad}F(\mathbf U))\oslash \mathbf U),\label{31March2014-2}
\end{align}
where $\text{grad}F(\mathbf U)$ is the Riemannian gradient and $D\text{grad}F(\mathbf U)[\xi_{\mathbf U}]$ is the Euclidean directional derivative of the Riemannian gradient in the direction $\xi_{\mathbf U}\in T_{\mathbf U}\mathbb{P}^K_M$, which is
\begin{align}
& D\text{grad}F(\mathbf U)[\xi_{\mathbf U}] =  D\Pi_{\mathbf U}(\text{Grad}F(\mathbf U)\odot \mathbf U)[\xi_{\mathbf U}] \notag\\
=& D\text{Grad}F(\mathbf U)[\xi_{\mathbf U}]\odot \mathbf U + \text{Grad}F(\mathbf U)\odot\xi_{\mathbf U} \notag\\
& - (\alpha \mathbf 1^T)\odot \xi_{\mathbf U} - (D\alpha[\xi_{\mathbf U}]\mathbf 1^T)\odot \mathbf U, \label{31March2014-3}
\end{align}
where $\alpha = (\text{Grad}F(\mathbf U)\odot \mathbf U)\mathbf 1$, $D\alpha[\xi_{\mathbf U}] = (D\text{Grad}F(\mathbf U)[\xi_{\mathbf U}]\mathbf U + \text{Grad}F(\mathbf U)\odot \xi_{\mathbf U})\mathbf 1$, and $D\text{Grad}F(\mathbf U)[\xi_{\mathbf U}]$ \changeBM{is} the Euclidean directional derivative of the Euclidean gradient $\text{Grad}F(\mathbf U)$ along the direction $\xi_{\mathbf U}\in T_{\mathbf U}\mathbb{P}^K_M$, \changeBM{i.e., $D\text{Grad}F(\mathbf U)[\xi_{\mathbf U}] = \lim\limits_{t \rightarrow 0} ( \text{Grad}F(\mathbf U + t \xi_{\mathbf U}) - \text{Grad}F(\mathbf U))/t$}.

\section{The Algorithm}\label{Sec:4}

The optimization problem \Gao{\eqref{30March2014-5} in Section \ref{Sec:2.3} is \changeBM{nonlinear}, with linear constraints. To this end, we propose to use the Riemannian trust-region (RTR) optimization algorithm \cite[Chapter~7]{AbsilMahonySepulchre2008}.}

\begin{table*}[htb]
\caption{\changeBM{Optimization-related ingredients for (\ref{30March2014-6}).}}
\begin{center}\small
\begin{tabular}{p{0.3\textwidth}|p{0.5\textwidth}}
\hline
& \\
Matrix representation of an element in the Multinomial Manifold $\mathbb{P}^K_M$ & A matrix $\mathbf U$ of size $M\times K$. \\

& \\
 $\mathbb{P}^K_M$ & $\mathbb{P}^K_M := \left\{\mathbf U: = [U_{mk}]\in\mathbb{R}^{M\times K}: \; U_{mk} > 0, \sum^{K}_{k=1}U_{mk} = 1\right\}$.\\

& \\
Tangent vectors in $T_{\mathbf U}\mathbb{P}^K_M$ & \changeBM{$\{ \xi_{\mathbf U} \in \mathbb{R}^{M\times K}:  \xi_{\mathbf U} \mathbf 1 = \mathbf 0\}$}, where $\mathbf 1\in\mathbb{R}^K$ is a column vector of ones.\\
& \\

Metric $g_{\mathbf U}(\xi_{\mathbf U}, \eta_{\mathbf U})$ for any $\xi_{\mathbf U}, \eta_{\mathbf U}\in T_{\mathbf U}\mathbb{P}^K_M$ & $g_{\mathbf U}(\xi_{\mathbf U}, \eta_{\mathbf U}) =\displaystyle \sum_{l,k}\frac{(\xi_{\mathbf U})_{mk}(\eta_{\mathbf U})_{mk}}{U_{mk}}$.\\
& \\

Projection of $\mathbf Z\in \mathbb{R}^{M\times K}$ onto the tangent space $T_{\mathbf U}\mathbb{P}^K_M$ with $\Pi_{\mathbf U}$& $\Pi_{\mathbf U}(\mathbf Z) := \mathbf Z - (\alpha \mathbf 1^T)\odot \mathbf U$, where $\alpha = \mathbf{Z1}^T$ and $\odot$ is the element-wise matrix multiplication operation.\\
& \\

Retraction $R_{\mathbf U}(\xi_{\mathbf U})$ that maps a search direction $\xi_{\mathbf U}$ onto $\mathbb{P}^K_M$ & $(\mathbf U \odot \exp (t (\xi_{\mathbf U} \oslash \mathbf U))) \oslash (\mathbf U \odot \exp (t (\xi_{\mathbf U} \oslash \mathbf U)) \mathbf 1\mathbf 1^T)$, where $\oslash$ is the element-wise matrix inversion and $\exp(\cdot)$ is the element-wise exponential operator.\\
& \\

\changeBM{Riemannian} gradient $\text{grad}F(\mathbf U)$ & $\Pi_{\mathbf U}(\text{Grad}F(\mathbf U)\odot \mathbf U)$, where $\text{Grad}F(\mathbf U)$ denotes the Euclidean gradient of function $F$.\\
& \\
\changeBM{Riemannian} Hessian $\text{Hess}F(\mathbf U)[\xi_{\mathbf U}]$ along $\xi_{\mathbf U} \in T_{\mathbf U}\mathbb{P}^K_M$ & $  \displaystyle \Pi_{\mathbf U}(D\text{grad}F(\mathbf U)[\xi_{\mathbf U}] - \frac12(\xi_{\mathbf U}\odot\text{grad}F(\mathbf U))\oslash \mathbf U)$, \changeBM{where $D\text{grad}F(\mathbf U)[\xi_{\mathbf U}]$ is defined in} (\ref{31March2014-3}).
\\
& \\
\hline
\end{tabular}
\label{TableIngredient}
\end{center}
\end{table*}

\subsection{The Riemannian Trust-Region Algorithm}\label{Sec:4.1}

\Gao{\changeBM{In order to} solve \eqref{30March2014-5}, \changeBM{we} define} $F(\mathbf U_N) = - \frac12\text{tr}(\mathbf B^T_N \mathbf U_N(\mathbf U^T_N\mathbf U_N)^{-1}\mathbf U^T_N \mathbf B_N)$. The problem \eqref{30March2014-5} \changeBM{boils down to} the optimization problem of the form
\begin{align}
\min_{\mathbf U\in \mathbb{P}^K_M} F(\mathbf U).\label{30March2014-6}
\end{align}

\changeBM{In order to simplify the exposition in the subsequent sections, we use $\mathbf U$ and $\mathbf B$ in (\ref{30March2014-6}) instead of $\mathbf U_N$ and $\mathbf B_N$, respectively. Similarly, we use $F$ instead of $-F$.}

The RTR algorithm is a generalization of the \changeBM{classical} unconstrained trust-region (TR) method to Riemannian manifolds. It is a matrix-free and globally convergent second-order method suitable for large-scale optimization
on Riemannian manifolds. Each iteration consists of two steps: (1) approximating the solution of the \emph{trust-region subproblem} and (2) computing a new iterate based on the retracting mapping, \changeBM{defined in Section \ref{sec:submanifold}}. The trust-region subproblem \changeBM{at $\mathbf{U} \in \mathbb{P}^K_M$} is defined as
\begin{align}
\min_{\xi_{\mathbf U}\in T_{\mathbf U}\mathbb{P}^K_M, \|\xi_{\mathbf U}\|\leq \Delta} F(\mathbf U) + g_{\mathbf U}(\text{grad}F(\mathbf U), \xi_{\mathbf U}) \notag\\ + \frac12 g_{\mathbf U}(\text{Hess}F(\mathbf U)[\xi_{\mathbf U}], \xi_{\mathbf U}),
\label{31March2014-4}
\end{align}
where $g$ is the Riemannian metric (\ref{1April2014-1}), $\Delta$ is \changeBM{the} trust-region radius, and $\|\xi_{\mathbf U}\| = \sqrt{g_{\mathbf U}(\xi_{\mathbf U}, \xi_{\mathbf U})}$. \eqref{31March2014-4} amounts to \changeBM{minimizing} a \emph{quadratic model} of the objective function within a trust-region radius of $\Delta$. \changeBM{Here, $\text{grad}F(\mathbf U)$ is the Riemannian gradient of $F$ and $\text{Hess}F(\mathbf U)[\xi_{\mathbf U}]$ is the Riemannian Hessian of $F$ along $\xi_{\mathbf U}$.}

The Riemannian gradient and Hessian of an objective function on the manifold can be computed \changeBM{using the expressions in} \eqref{31March2014-1}, \eqref{31March2014-2} and \eqref{31March2014-3} from \changeBM{the} Euclidean gradient $\text{Grad}F(\mathbf U)$ and Euclidean Hessian $D\text{Grad}F(\mathbf U)[\xi_{\mathbf U}]$ counterparts.

For our objective function
\[
F(\mathbf U) = - \frac12\text{tr}(\mathbf B^T \mathbf U(\mathbf U^T\mathbf U)^{-1}\mathbf U^T \mathbf B),
\]
it is \changeBM{straightforward} to check, using matrix calculus \cite{PetersenPetersen2008}, that
\begin{align*}
& \text{Grad}F(\mathbf U) \\
 = & - \mathbf B\mathbf B^T\mathbf U(\mathbf U^T\mathbf U)^{-1} + \mathbf U(\mathbf U^T\mathbf U)^{-1}\mathbf U^T\mathbf B\mathbf B^T\mathbf U(\mathbf U^T\mathbf U)^{-1}
\end{align*}
and
\begin{align*}
& D\text{Grad}F(\mathbf U)[\xi_{\mathbf U}] \\
= & - \mathbf B\mathbf B^T\xi_{\mathbf U} (\mathbf U^T\mathbf U)^{-1} + \xi_{\mathbf U} (\mathbf U^T\mathbf U)^{-1}(\mathbf U^T\mathbf B\mathbf B^T\mathbf U)(\mathbf U^T\mathbf U)^{-1}\\
 & +  2 \mathbf B\mathbf B^T\mathbf U(\mathbf U^T\mathbf U)^{-1}\text{sym}(\xi^T_{\mathbf U}\mathbf U)(\mathbf U^T\mathbf U)^{-1}\\
 & + 2\mathbf U(\mathbf U^T\mathbf U)^{-1}\text{sym}(\xi^T_{\mathbf U}\mathbf B\mathbf B^T\mathbf U)(\mathbf U^T\mathbf U)^{-1}\\
 -&2\mathbf U(\mathbf U^T\mathbf U)^{-1}\text{sym}(\mathbf U^T\xi_{\mathbf U})(\mathbf U^T\mathbf U)^{-1}\mathbf U^T\mathbf B\mathbf B^T\mathbf U(\mathbf U^T\mathbf U)^{-1}\\
 -&2\mathbf U(\mathbf U^T\mathbf U)^{-1}\mathbf U^T\mathbf B\mathbf B^T\mathbf U(\mathbf U^T\mathbf U)^{-1}\text{sym}(\mathbf U^T\xi_{\mathbf U})(\mathbf U^T\mathbf U)^{-1},
\end{align*}
where $\text{sym}{(\mathbf A)} = (\mathbf A+\mathbf A^T)/2$ \changeBM{extracts the symmetric part of $\mathbf{A}$}.

Once the expressions of the Riemannian gradient and Hessian are obtained, we use the Riemannian trust-region (RTR) algorithm implemented in the Manopt toolbox \cite{BoumalMishraAbsilSepulchre2014}. \Gao{With all the ingredients summarized in Table \ref{TableIngredient}, the main steps of the RTR algorithm \changeBM{are} presented in Fig. \ref{fig:RTR} for the paper to be self-contained.}

\begin{figure}[h]
\hrule\hrule\vspace{5pt}
\begin{algorithmic}[1]
\REQUIRE An initial guess $\mathbf U_0$ on the manifold $\mathbb{P}^{K}_M$. \\
\ENSURE  The minimum $\mathbf U$ for the objective function $F$.\\
\STATE Continue the following for loop until a convergence criterion is satisfied.
   \FOR {$i=1, 2, ...$}
    \STATE Approximately minimize the trust-region subproblem \eqref{31March2014-4} for a new direction $\xi_{\mathbf U}$.
    \STATE Construct the new trial iterate by using retraction mapping $\mathbf U_+=R_{\mathbf U}(\xi_{\mathbf U})$. \\
    \STATE Update the iterate by rejecting or accepting $\mathbf U_+$ depending on its quality \changeBM{of decrease in the objective function}.
    \STATE Update the trust-region radius $\Delta$.
   \ENDFOR
\end{algorithmic}
\vspace{5pt}
\hrule
\caption{The Riemannian trust-region algorithm.}
\label{fig:RTR}
\end{figure}

\subsection{Overall Algorithm}\label{OverallAlgorithm}

The overall algorithm for (\ref{30March2014-4}) is based on an alternating optimization scheme in which one matrix variable is updated while \changeBM{fixing the} other matrix variables. In Section \ref{Sec:2.3}, the updates of the matrix variables $\mathbf U_1, \mathbf U_2, ..., \mathbf U_{N-1}$ are shown in closed form. The update of the last mode $\mathbf U_N$ is computed by solving (\ref{30March2014-5}) with \changeBM{the} Riemannian trust-region algorithm discussed in Section \ref{Sec:4.1}. The procedure of cyclically updating the variables is repeated until a convergence criterion is satisfied. Finally, \changeBM{the} membership information $\mathbf U_N$ is used with a clustering algorithm, such as the k-means, \changeBM{to learn the final clustering parameters}.

The overall algorithm is  summarized in Fig. \ref{fig:overall_algorithm}.

\begin{figure}[h]
\hrule\hrule\vspace{5pt}
\begin{algorithmic}[1]
\REQUIRE Tensorial data $\mathcal{X}=\{\mathcal{X}_1, ..., \mathcal{X}_M\}$, dimensions $I_1, I_2, \ldots, I_{N-1}$ and the number of clusters $K$. \\
\ENSURE  Factor Matrices $\mathbf U_1, \mathbf U_2, ..., \mathbf U_{N-1}$ and Clusters.\\
   \STATE Initialize $\mathbf U_1, \mathbf U_2, ..., \mathbf U_{N-1}$.
   \STATE Continue the following for loop until a convergence criterion is satisfied.
   \FOR {$i=1, 2, ...$}
    \STATE Call the Riemannian trust-region algorithm (RTR) in Fig. \ref{fig:RTR} to compute $\mathbf U_N$.
    \FOR {$n=1, 2, ..., N-1$}
        \STATE Solve \eqref{31March2014-5} for $\mathbf U_n$ by using  \eqref{31March2014-6}.
    \ENDFOR
   \ENDFOR
   \STATE Do k-means over $\mathbf U_N$ for the final clustering.
\end{algorithmic}
\vspace{5pt}
\hrule
\caption{The overall algorithm for (\ref{30March2014-4}).}
\label{fig:overall_algorithm}
\end{figure}

\textit{Remark 1:} The tensor clustering model and algorithm, Adaptive
Subspace Iteration on Tensor (AST-T), proposed in \cite{PengLi2011a} is a special version of \changeBM{the} HOSVD model. The idea is to combine tensor projection and k-means by minimizing the distance of projected tensors and the one of $K$ tensor centroids. The objective function is different from our formulation. AST-T uses a heuristic method to approximately solve the relevant optimization problem.

\textit{Remark 2:} Our model is also different from the fuzzy k-means algorithm in which the fuzzy cluster membership parameters are applied over the distance between data and centroids. In our model, we regard each data as a linear combination of centroids under relevant membership coefficients.

\textit{Remark 3:} The algorithm in Fig. \ref{fig:overall_algorithm} starts the \textit{for} loop with updating the membership information $\mathbf U_N$. We can also start the for loop with updating factor matrix $\mathbf U_1, \mathbf U_2, ..., \mathbf U_{N-1}$. In this case, we can use k-means to make an estimate for $\mathbf U_N$.

\textit{Remark 4:} \changeBM{The overall algorithm is an alternating optimization scheme for (\ref{30March2014-4}). The general convergence analysis of alternating optimization schemes is discussed in \cite[Section~8.9]{Luenberger2008}. The convergence analysis of RTR is discussed in \cite[Chapter~7]{AbsilMahonySepulchre2008}. It should be emphasized that the objective function in (\ref{30March2014-5}) is nonlinear and RTR, in general, converges to a critical point or a local minimum. Nonetheless, this suffices for the scheme in Fig. \ref{fig:overall_algorithm} for solving (\ref{30March2014-4}), which is a challenging optimization problem.}

\subsection{Initialization}\label{sec:initialization}
Since we have a nonlinear objective function to be optimized over the ``curved'' multinomial manifold, an initialization has \changeBM{an} impact on the final result. In our experiments, \changeBM{shown} in the next section, we consider\changeBM{ the following} strategies for initialization.

\textit{Random Initialization:}  This suggests using randomly produced orthogonal basis as each of \changeBM{the} factor matrices $\mathbf U_1, \mathbf U_2, ..., \mathbf U_{N-1}$.

\textit{HOSVD Initialization I:} Given the data tensor $\mathcal{X}$, we conduct the HOSVD or HOOI. \changeBM{Keeping the orthogonal matrix factors $\mathbf U_1, \mathbf U_2, ..., \mathbf U_{N-1}$ fixed to the initial values, we update $\mathbf U_N$ by using the Riemannian trust-region algorithm. However, the orthogonal matrix factors generated by the HOSVD algorithm may not accurately represent data for clustering. In other words, the initialization for $\mathbf U_N$ from the HOSVD is far from a good membership representation of clusters. As a result, the the first call to the RTR algorithm requires a larger number of iterations, e.g., $1000$.}

\textit{HOSVD Initialization II:} The HOSVD or HOOI decomposition is designed for a single tensor. As we have to single out the last mode for the purpose of clustering, we design a similar HOSVD algorithm for the first $(N-1)$ \changeBM{modes}. In other words, \changeBM{the resulting problem is over a} group of $(N-1)$-order tensors $\{\mathcal{X}_i\}^M_{i=1}$ for which we learn their Tucker decompositions under a fixed set of factor matrices $\{\mathbf U_1, ..., \mathbf U_{N-1}\}$ such that the optimization problem
\[
\sum^M_{i=1} \|\mathcal{X}_i - \mathcal{G}_i\times_1 \mathbf U_1 \times_2 \cdots \times_{N-1}\mathbf U_{N-1}\|^2_F
\]
is minimized. This problem can be solved in a way similar to the problem of computing the single tensor HOSVD. However, this initialization may not provide any information relating to the tensor representation in terms of the $K$ centroids along the $N$-th mode.

\section{Numerical Experiments}\label{Sec:5}

\changeBM{In this section, we present a set of experimental results on synthetic and real-world datasets with high dimensional spatial structures. The intention of these experiments is to demonstrate the performance of our proposed model in comparison to a number of state-of-the-art clustering methods. The algorithm proposed in Fig. \ref{fig:overall_algorithm} compete effectively with the benchmark methods in terms of prediction accuracy.}

\subsection{Evaluation Metrics}
To quantitatively evaluate the clustering results, we adopt two evaluation metrics, the \emph{accuracy} (AC) and the \emph{normalized mutual information} (NMI) \changeBM{metrics} \cite{CaiHeWangBaoHan2009}. Given a data point $\mathbf{x}_i$, let $L$ and $\hat{L}$ be the ground truth label and the cluster label provided by the clustering approaches, respectively. \changeBM{The AC} measure is defined by
\[
\textrm{AC} = \frac{{\sum\limits_{i = 1}^M {\delta \left( {\hat{L}(i),\textrm{Map}_{{(\hat{L},L)}}(i)} \right)} }}{M},
\]
where $M$ is the \changeBM{total} number of samples and \changeBM{the} function $\delta ( \mathbf{a}, \mathbf{b} )$ is set to 1 if and only if $\mathbf{a} =\mathbf{b}$, and $0$ otherwise. \changeBM{The operator} Map$( \cdot )$ is the \emph{best mapping function} that permutes $\hat{L}$ to match $L$, which is usually implemented by the Kuhn-Munkres algorithm \cite{ChenDonohoSaunders2001}.

The other metric is the normalized mutual information \changeBM{measure} between two index sets $L$ and $\hat{L}$, defined as
\[
\textrm{NMI}(L,\hat{L}) =  \frac{\textrm{MI}(L,\hat{L})}{\max\left(\textrm{H}\left(L\right),\textrm{H}\left(\hat{L}\right)\right)},
\]
where  $\textrm{H}(L)$ and $ \textrm{H}(\hat{L})$ denote the entropy of  $L$  and  $\hat{L}$, respectively, and
\[
\textrm{MI}\left(L, \hat{L}\right) = \sum\limits_{y \in L} {\sum\limits_{x \in {\hat{L}}} {p\left( {x,y} \right){{\log }_2}\left( {\frac{{p\left( {x,y} \right)}}{{p\left( x \right)p\left( y \right)}}} \right)} }.
\]
$p\left( y \right)$ and $p\left( x \right)$ denote the marginal probability distribution functions of $L$ and $\hat{L}$, respectively, and $p\left( {x,y} \right)$ is the joint probability distribution function of $L$ and $\hat{L}$. $\textrm{NMI}(L,\hat{L})$ ranges from $0$ to $1$, for which the value $1$ means that the two sets of clusters are identical and the value 0 means that the two are independent. Different from AC, NMI is \emph{invariant} \changeBM{to} permutation of labels, i.e., it does not require the matching processing in advance.

\subsection{Dataset Description}
This section \Gao{describes the real-world datasets} that we use for assessing the \changeBM{performance of various clustering algorithms}.

\textit{CBCL face dataset}: \changeBM{This face dataset\footnote{http://cbcl.mit.edu/projects/cbcl/software-datasets/FaceData2. html} contains images of size $19\times 19$}. The goal of clustering for this dataset is to cluster the images into two different classes: face and non-face.

\textit{MNIST dataset}: This  handwritten digits dataset\footnote{http://yann.lecun.com/exdb/mnist/} consists of $70,000$ handwritten digit images, including a training set of $60,000$ examples and a test set of $10,000$ examples. It is a subset extracted from a larger set available from NIST. The digits have been size-normalized and centered in the fixed-size $28\times 28$. The images are in grey scale and each image can be treated as a $784$-dimension feature vector or a $28\times 28$ second order tensor. This dataset has $10$ classes corresponding to the digits \changeBM{0 to 9}, with all \changeBM{the} images being labeled.

\textit{PIE dataset, CMU}: The CMU Pose, Illumination, and Expression (PIE) dataset\footnote{http://www.ri.cmu.edu/research\_project\_detail.html?project\_id =418\&menu\_id=261} consists of $41,368$ images of $68$ people. Each subject was imaged under $13$ different poses, $43$ different illumination conditions, and with $4$ different expressions. In this paper, we test the algorithms (ours and the benchmarks) on the \emph{Pose27 sub-dataset} as described in \cite{CaiHeHanHuang2011}. The image size is $32\times 32$.

\textit{ORL dataset}:
The AT\&T ORL dataset\footnote{http://www.uk.research.att.com/facedatabase.html.} consists of 10 different images for each of 40 distinct subjects, thus 400 images in total. All the images were taken against a dark homogeneous background with the subjects in an upright, frontal position, under varying lighting, facial expressions (open/closed eyes, smiling/not smiling), and facial details (glasses/no glasses). \changeBM{For our experiments}, each image is resized \changeBM{to} $32\times 32$ pixels.

\textit{Extended Yale B dataset}: For this dataset\footnote{http://vision.ucsd.edu/\~{}leekc/ExtYaleDatabase/ExtYaleB.html}, we use the cropped images and resize them to $32\times 32$ pixels. This dataset now has 38 individuals and around 64 near frontal images under different illuminations per individual.


\Gao{\textit{Dynamic texture dataset:} The DynTex++ dataset\footnote{http://www.bernardghanem.com/datasets} consists of video sequences that contains dynamic river water, fish swimming, smoke, cloud and so on. These videos are labeled with $36$ classes and each class has 100 subsequences (a total of $3600$ subsequences) with a fixed size of $50\times 50\times 50$ ($50$ gray frames). This is a challenging dataset for clustering because most of texture from different class is fairly similar.}

\subsection{Assessing Initialization Strategies}
In Section \ref{sec:initialization} we \Gao{propose} three different initialization schemes for the proposed tensor clustering algorithm. In this experiment, we assess the influence of \changeBM{those} initialization \changeBM{strategies on} the final clustering accuracy on the sub-dataset Pose27 from the PIE dataset.

Before we present the experimental results, we describe \changeBM{the} parameter settings used in the experiments. We randomly select 1000 data from the sub-dataset Pose27. There are $24$ different classes in this chosen subset. The image size is $32 \times 32$. We choose the size of core tensors to be $12 \times 12$. In the overall algorithm described in Fig. \ref{fig:overall_algorithm}, \changeBM{we fix the number of iterations to $250$ as the recovered error does not vary much after $250$ iterations in most testing cases. For the RTR algorithm in line $4$ of Fig. \ref{fig:overall_algorithm}, we use the default parameters proposed in the toolbox Manopt with the RTR maximal inner iteration number $30$. For the first call to the RTR algorithm, we set the number of maximum outer iterations to $1000$ with  Manopt's default initialization for $\mathbf U_N$ ($N=3$ in this case). For subsequent calls to RTR, we use $5$ iterations with the current $\mathbf U_N$ as the initial values for memberships.} In each overall iteration, we repeat the calls to Lines $5$-$7$ in Fig. \ref{fig:overall_algorithm} twice to make sure that the factors $\mathbf U_n$ stable.

\begin{table}[t]
\caption{Results for random Initialization.}\label{Table1}
\begin{center}
\begin{tabular}{|c|c|c|c|c|}
\hline
\multicolumn{1}{|c|}{Evaluation metric}%
& \multicolumn{1}{c|}{Mean}%
& \multicolumn{1}{c|}{Std Var}%
& \multicolumn{1}{c|}{Best}
& \multicolumn{1}{c|}{Worst}\\%
\hline
AC  & 0.7120  & 0.0394  &  0.7630  & 0.6150 \\
NMI & 0.8189  & 0.0210 & 0.8467 & 0.7717\\ \hline
\end{tabular}
\label{Table1}
\end{center}
\end{table}

\begin{figure}
    \begin{center}
     \subfloat[HOSVD Initialization I]{\includegraphics[width=0.35\textwidth]{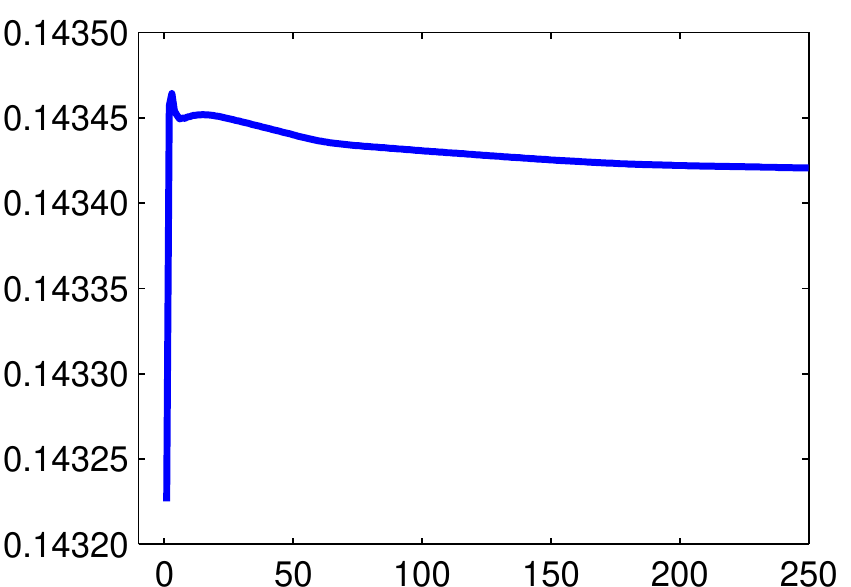}} \\
     \subfloat[HOSVD Initialization II]{\includegraphics[width=0.35\textwidth]{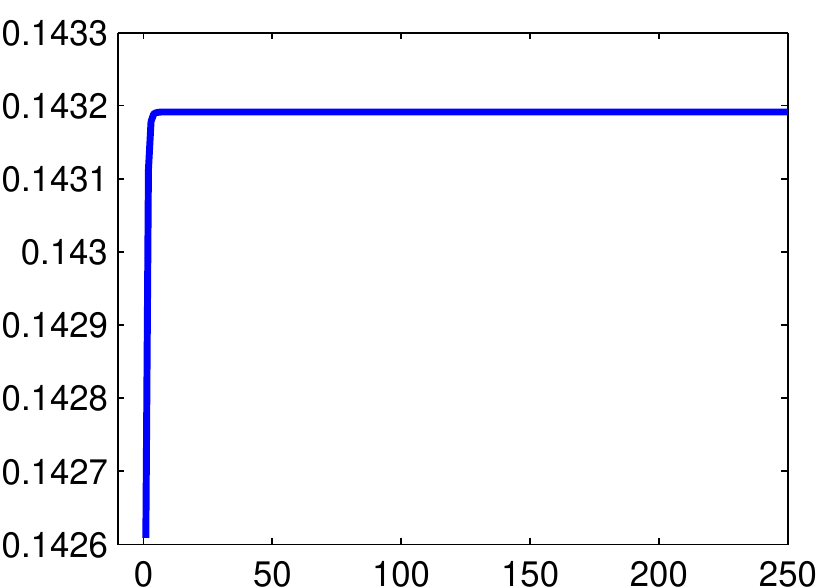}}
     \caption{The errors against the number of iterations.}\label{Figure3}
    \end{center}
\end{figure}

\changeBM{We \Gao{conduct} 20 runs for randomly initializing $\mathbf U_1$ and $\mathbf U_2$. The statistics are reported in Table \ref{Table1}. Under similar parameter settings, $\text{AC} = 0.7620$ and $\text{NMI}=0.8199$ for HOSVD Initialization I and $\text{AC}= 0.7240$ and $\text{NMI} = 0.8132$ for HOSVD Initialization II, respectively. Both the random initialization and HOSVD initialization II strategies give comparable results and are are shown in Fig. \ref{Figure3}(b). It should be noted that the reconstructed error does not improve with iterations. On the other hand, Fig. \ref{Figure3}(a) shows the error curve for the HOSVD Initialization I strategy.  As HOSVD gives the best estimate in terms of the model error, the error goes up initially when we move away from the best orthogonal factor $\mathbf U_3$ to a membership factor. Finally, it stabilizes as shown in Fig. \ref{Figure3}(a). This leads to the conclusion that among the three initialization strategies discussed in Section \ref{sec:initialization}, HOSVD initialization I is the initialization of choice. In subsequent experiments, we initialize all the compared algorithms with HOSVD initialization I. }

\subsection{Algorithms and Comparisons}
There exist many different clustering algorithms such as \emph{spectral clustering approaches} \cite{ShiMalik2000} and the recent popular \emph{matrix factorization} approaches\cite{LeeSeung2001,KimPark2007}. Since the Tucker decomposition is a multidimensional generalization of matrix factorization, we focus on the comparison between our proposed heterogeneous Tucker model with recent matrix factorization based clustering algorithms. The following methods are chosen as the benchmark for numerical comparisons. \Gao{All the experiments are conducted on a desktop with an Intel Core i5-4670 CPU at 3.40GHz and with RAM of 8.00GB.}

\textit{Multiplicative method for nonnegative matrix factorization (MM) \cite{LeeSeung2001}:} \changeBM{This nonnegative matrix factorization algorithm is designed for vectorial data that are organized into a data matrix $\mathbf X$, i.e., each column represents an image. The matrix $\mathbf X$ is factorized into the product of two nonnegative matrices such that $\mathbf X = \mathbf{WH}$. The columns of $\mathbf W$ are regarded as the \emph{basis vectors} and $\mathbf H$ is interpreted as the \emph{weights}, i.e.,  each data vector is obtained by a weighted linear combination of the the basis vectors. The matrix $\mathbf H$ is also interpreted as containing clustering membership information.}

\textit{Alternating least-squares algorithm (ALS) \cite{KimPark2007}:} This algorithm \Gao{is} designed to solve the nonnegative matrix factorization problem of factorizing the nonnegative data matrix $\mathbf X$ as $ \mathbf{WH}$ by \emph{alternating optimization} over the factor matrices $\mathbf W$ and $\mathbf H$ with the nonnegativity constraint.

\textit{Bayesian nonnegative matrix factorization (B-NFM) \cite{SchmidtWintherHansen2009}:} The model is based on the probabilistic framework in which the matrix factors $\mathbf W$ and $\mathbf H$ are regarded as random variables characterized by relevant statistical density functions. A \emph{Bayesian algorithm} \Gao{is} proposed for solving the problem.

\textit{Projected gradient method for nonnegative matrix factorization (PGM) \cite{Lin2007}:} This algorithm results from simultaneously optimizing over the nonnegative factors $\mathbf W$ and $\mathbf H$ to approximately decompose a nonnegative data matrix $\mathbf X$ as $ \mathbf{WH}$. Lin \cite{Lin2007} \Gao{proposes} an efficient \emph{Newton} algorithm.

\textit{Dual regularized co-clustering method (DRCC) \cite{GuZhou2009}:} The DRCC algorithm is a data clustering algorithm that is based on \emph{semi-nonnegative matrix tri-factorization} with dual regularization over data graph and feature graph by exploiting the geometric structure of the data manifold and the feature manifold.

\textit{Nonnegative tri-factor tensor decomposition (Tri-ONTD) \cite{ZhangLiDing2013}:} This algorithm is designed for 3-order tensorial data. It proposes a tensor decomposition with nonnegative factor matrices. Since it directly deals with tensor data, it can be considered as a benchmark algorithm for comparison with our proposed method.

\textit{Benchmark PCA-B:} To demonstrate the benefits of the proposed model and, particularly, the Riemannian algorithm, we take the following revised algorithm as another benchmark algorithm. Instead of the problem \eqref{30March2014-5}, we propose to consider the problem \eqref{30March2014-1} for a nonnegative membership matrix $\mathbf U_N$ only, i.e., the constraint on the sum of each row to $1$ is lifted. This makes the problem formulation simpler, which can be solved by alternatively optimizing of each factor matrices, similar to the procedure in Fig. \ref{fig:overall_algorithm}. In this case, optimizing $\mathbf U_N$ becomes a \emph{quadratic programming problem with nonnegative constraints}, which can be solved, e.g., using the Matlab function \texttt{lsqlin.m}.

\changeBM{\textit{PCA-Multinomial:} The algorithm proposed in Fig. \ref{fig:overall_algorithm}, which is based on the proposed heterogeneous Tucker decomposition model discussed in Section \ref{Sec:2.2}.}

\changeBM{It should be noted that most above described algorithms perform better than (or comparable in some cases) to classical clustering algorithms, like the k-means \cite{GuZhou2009,ZhangLiDing2013}. Hence, we do not compare with the classical clustering algorithms for vectorized data.}

\Gao{The implementations of the nonnegative matrix factorization algorithms MM, ALS, B-NFM, and PGM are in the Matlab NMF toolbox provided by Kasper Winther Joergensen.} The Matlab code for DRCC is provided by the authors of \cite{GuZhou2009}. We \Gao{implement} all the other algorithms. \changeBM{Our proposed algorithm, PCA-Multinomial, is implemented in Matlab using the Manopt toolbox \cite{BoumalMishraAbsilSepulchre2014}.}

\subsubsection*{Experiment I}
We first \Gao{test} all the algorithms on the CBCL face dataset as this is the simplest case where there are only two clusters of images \changeBM{to be learned}. The datasizes are chosen from $200$ to $1600$ by $100$ for each class and the experiment results indicate that when size $=600$ almost all the algorithms achieved similar accuracy.  Under this size, we also \Gao{test} different tensor core sizes  \changeBM{varying} from $4$ to $10$ and \Gao{find} better results for our proposed tensor clustering method in the case of tensor core size $8$.

\changeBM{Table \ref{Table2} shows the means and variances (in brackets) for all the algorithms after $10$ runs. In Table \ref{Table2} we see that all the algorithms are comparable for the case when the number of clusters is $2$. For this experiment, DRCC has shown better performance than all the other algorithms. However, we also \Gao{observe} that in experiments when $K = 2$ the performance of DRCC degrades. The benchmark algorithm PCA-B is slightly better than the proposed PCA-Multinomial by $1.8\%$ in AC. However, the variance of PCA-Multinomial is better in this experiment showing robustness of the Riemannian optimization algorithm.}
\begin{table*}[htb]
\caption{Results for the CBCL face dataset with 1200 randomly chosen data (600 from each class).}
\begin{center}
\begin{tabular}{|c|c|c|c|c|c|c|c||c|}
\hline
\multicolumn{1}{|c|}{{Evaluation metric}}%
& \multicolumn{1}{c|}{MM}%
& \multicolumn{1}{c|}{ALS}%
& \multicolumn{1}{c|}{B-NFM}
& \multicolumn{1}{c|}{PGM}
& \multicolumn{1}{c|}{DRCC}
& \multicolumn{1}{c|}{Tri-ONTD}
& \multicolumn{1}{c||}{PCA-B}
& \multicolumn{1}{c|}{PCA-Mulitnomial}\\%
 \hline
AC & 0.6755  & 0.6424  & 0.6305 & 0.5976 & \textbf{0.7352} & 0.5764  &0.6394 & {0.6208 }\\
   & (5.494e-4)  & (3.255e-4)  & (3.594e-4) & (2.073e-3) & (1.367e-3) & (3.503e-5) &(3.301e-3) & (1.018e-4)\\
NMI & 0.0928   & 0.0604  & 0.0509  & 0.0334 & \textbf{0.1766} & 0.0824  &0.0669 & {0.0430} \\
  & (6.351e-4)  & (2.505e-4) & (2.296e-4) & (7.794e-4) & (2.777e-3) & (1.836e-4) &(1.181e-3)& (5.524e-5) \\ \hline
\end{tabular}
\label{Table2}
\end{center}
\end{table*}

\subsubsection*{Experiment II}
In this experiment, we assess the impact of the number of classes on the performance of all the algorithms. For this purpose, we conduct all the algorithms on the MNIST handwritten digits dataset. There are ten different classes, corresponding to the digits $0$ to $9$. Most machine learning algorithms work well for the case of two clusters, i.e., $K = 2$. For $K > 2$, the performance of different algorithms varies, observing which is the main focus of the present experiment.

We randomly \Gao{pick} digits for training data according to the class number $K=3, 4, ..., 10$ with $100$ images for each class. Once the classes are decided randomly, we run all the algorithms $5$ times each with randomly chosen digits in the classes. The experimental results are reported in Table \ref{Table3} in terms of AC and NMI evaluation metrics. \Gao{It should be noted that} Tri-ONTD algorithm fails for most of cases, and therefore, no results are reported. The results in Table \ref{Table3} show that the proposed PCA-Mulitnomial performs better in the cases of more than four classes. This is consistent with the results from the first experiment, where we observe a similar behavior. For $K =3, 4$, the performance of PCA-Multinomial is comparable with that of MM and PGM in both AC and NMI metrics.

\begin{figure}
\begin{center}
 \includegraphics[width=0.12\textwidth]{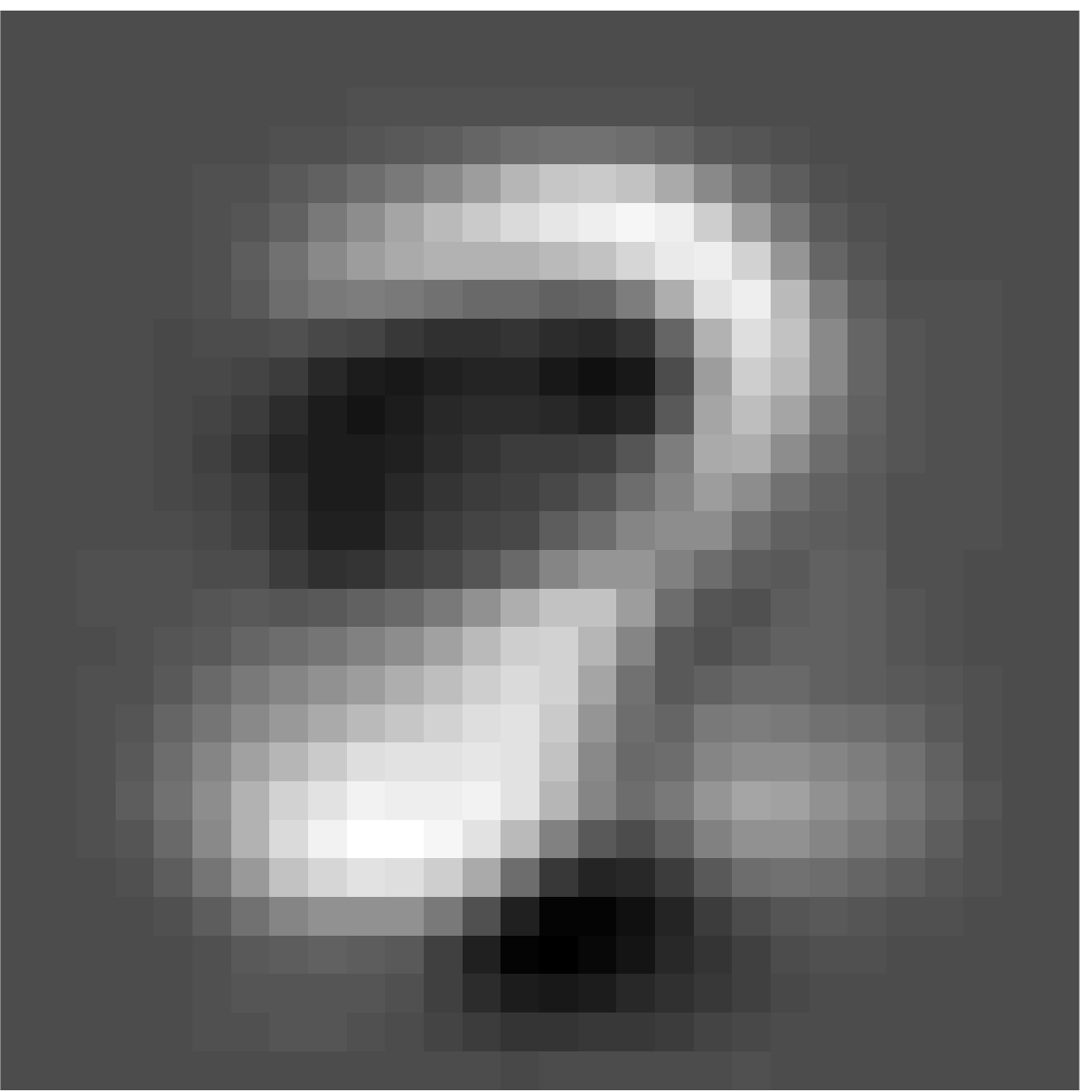}\includegraphics[width=0.12\textwidth]{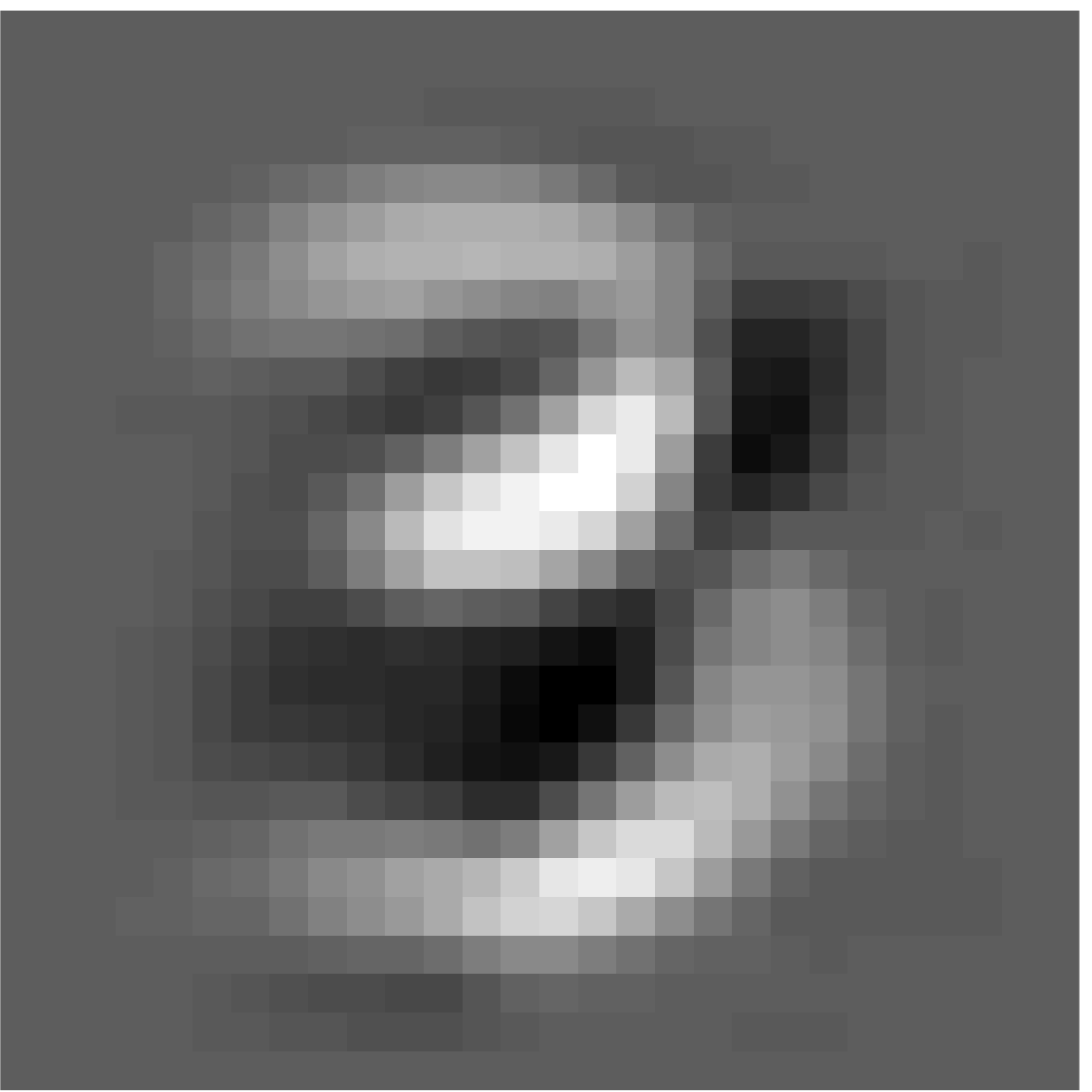}\includegraphics[width=0.12\textwidth]{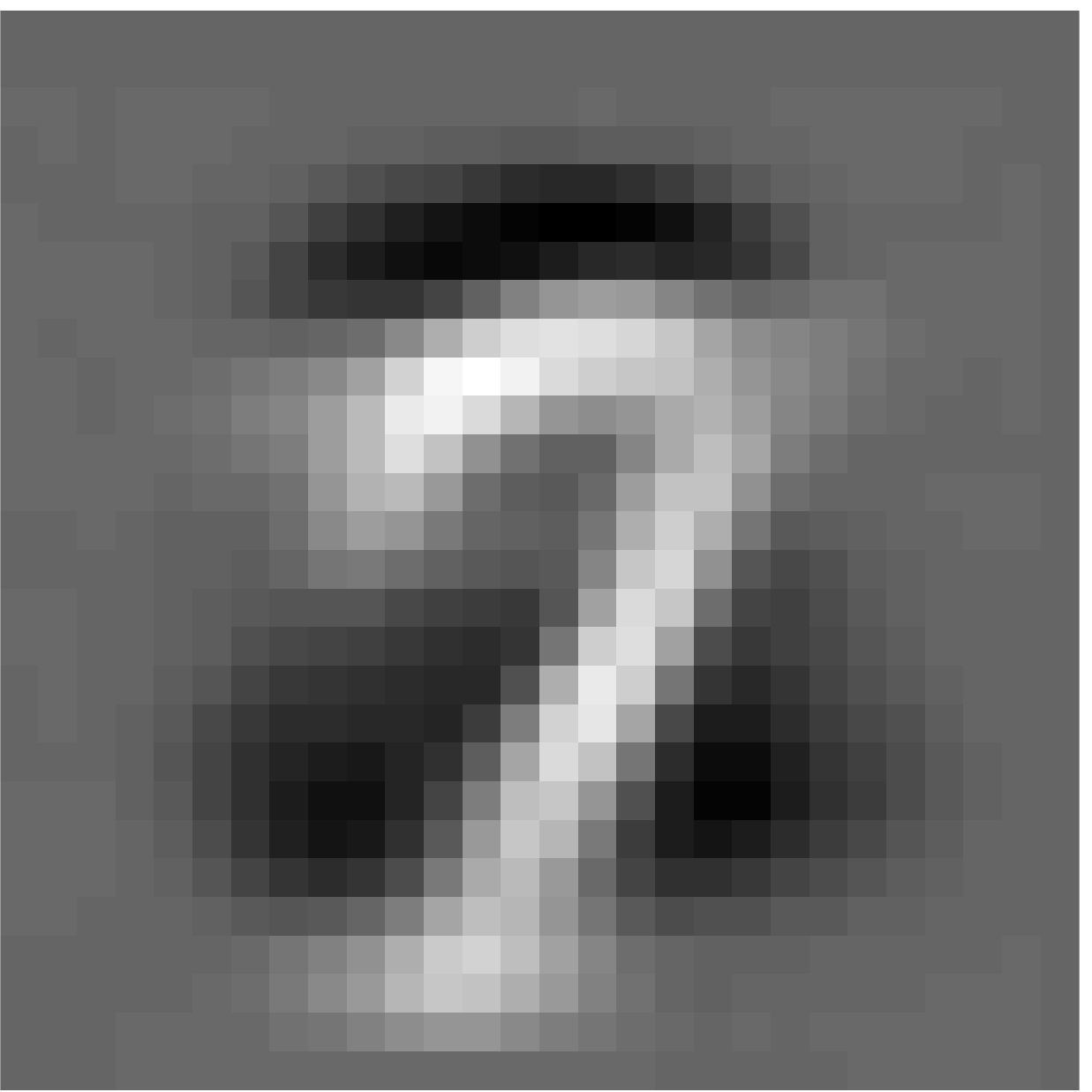}\includegraphics[width=0.12\textwidth]{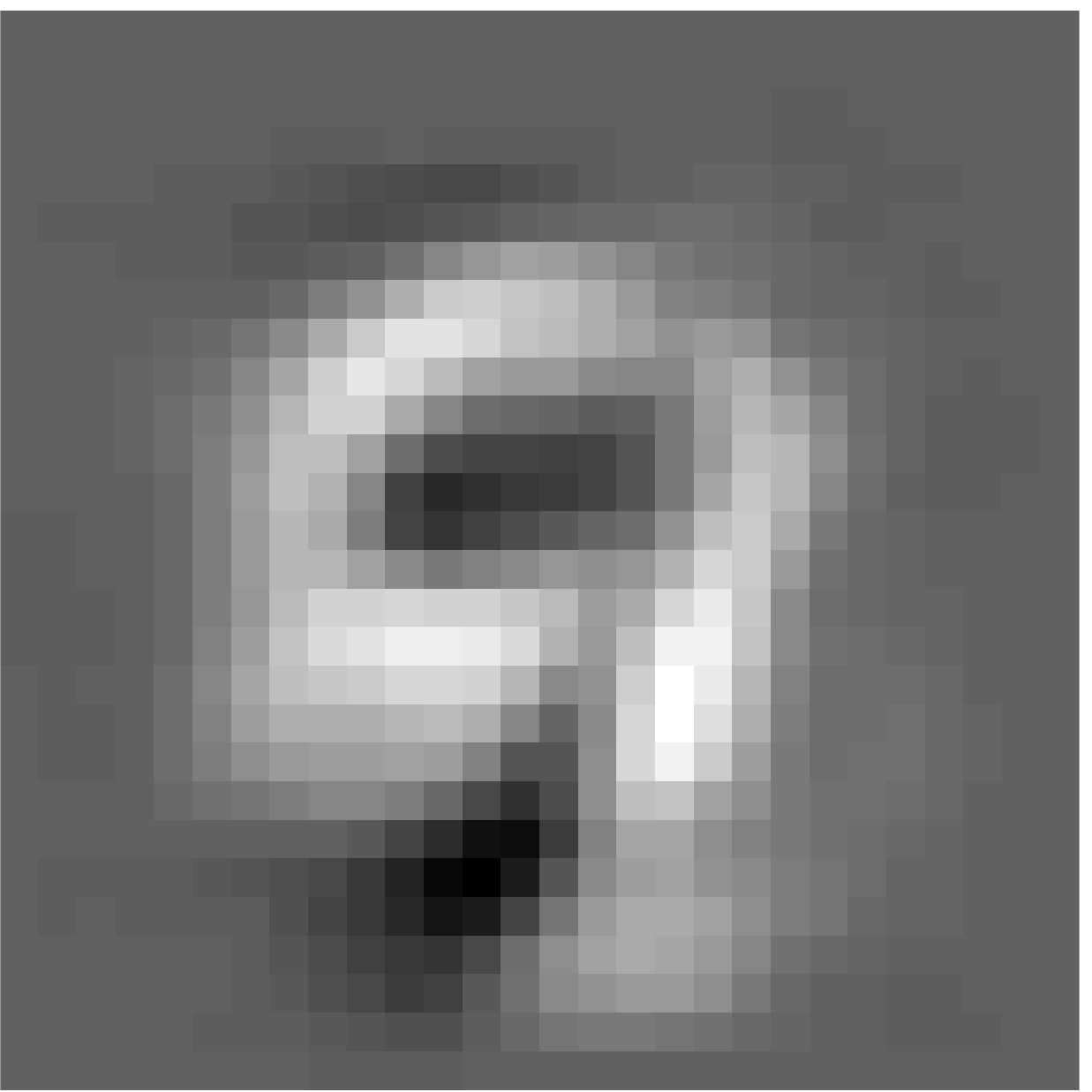}\\
 \includegraphics[width=0.12\textwidth]{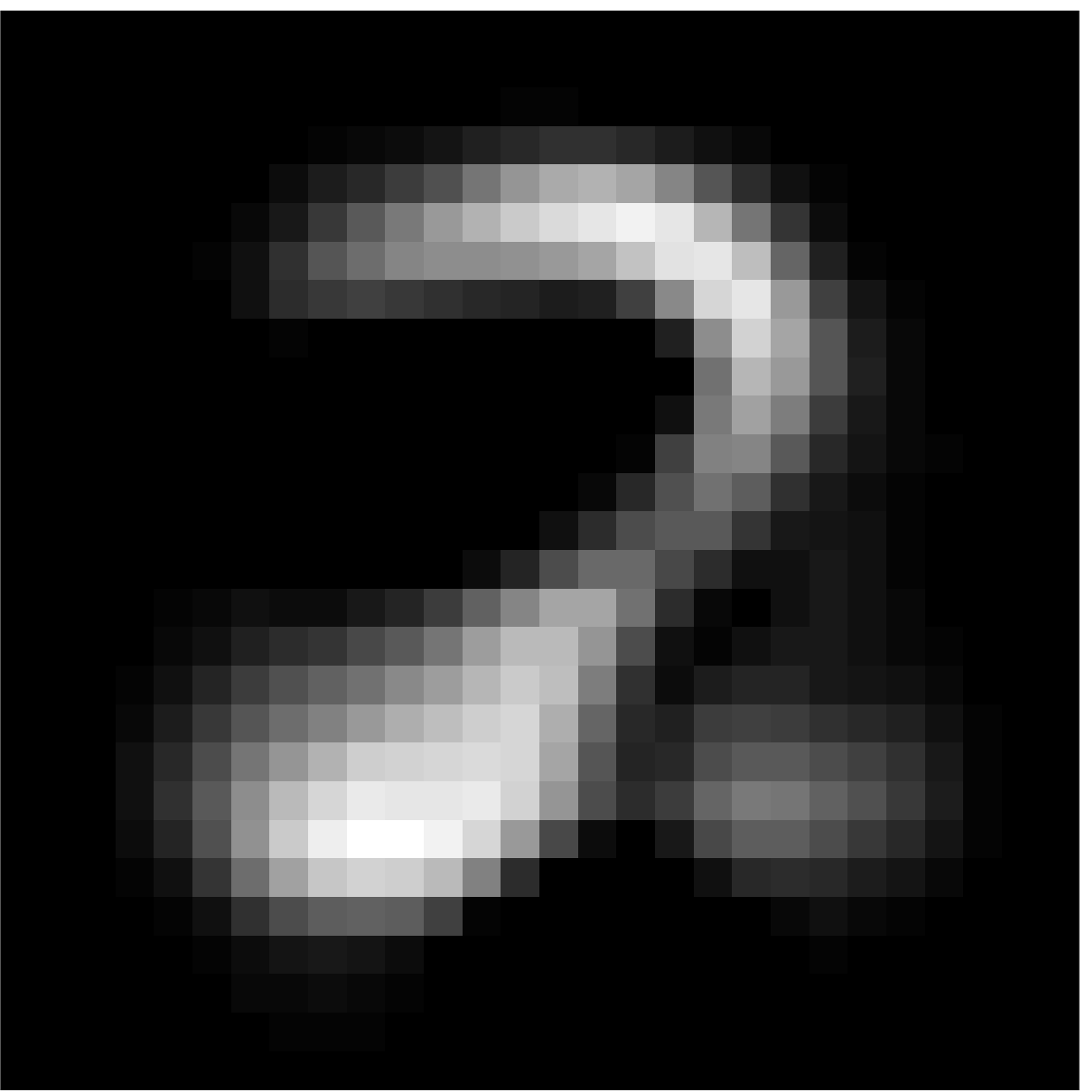}\includegraphics[width=0.12\textwidth]{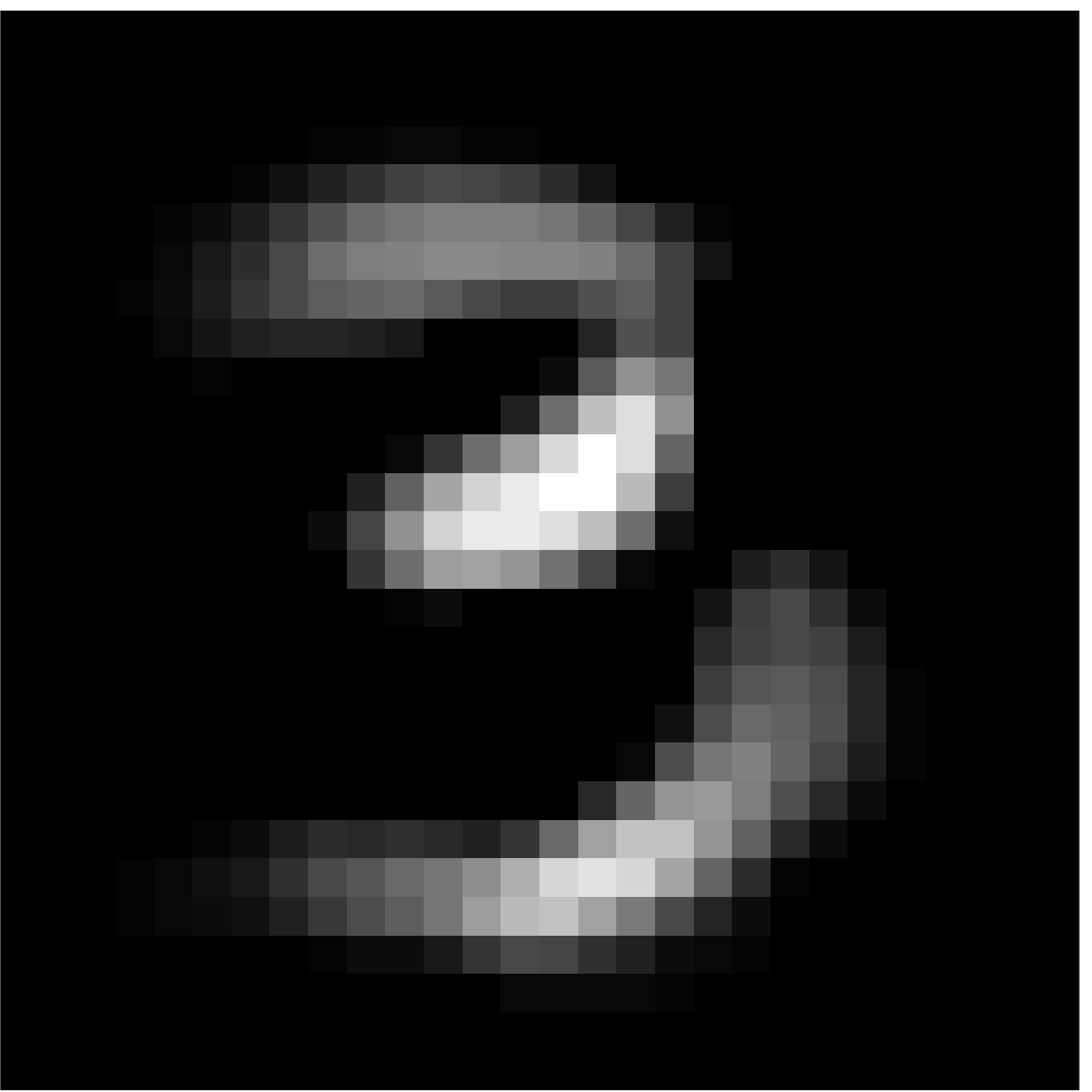}\includegraphics[width=0.12\textwidth]{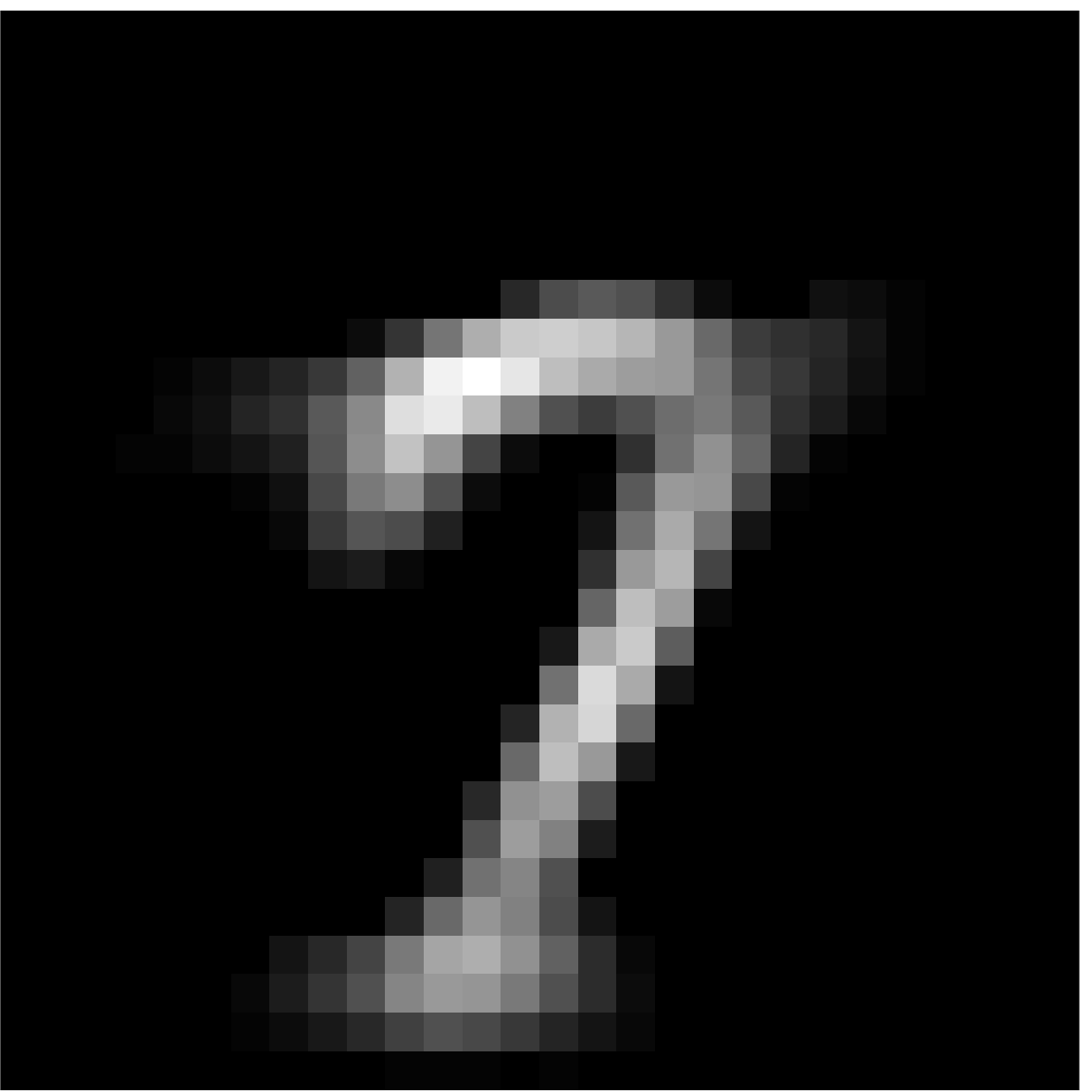}\includegraphics[width=0.12\textwidth]{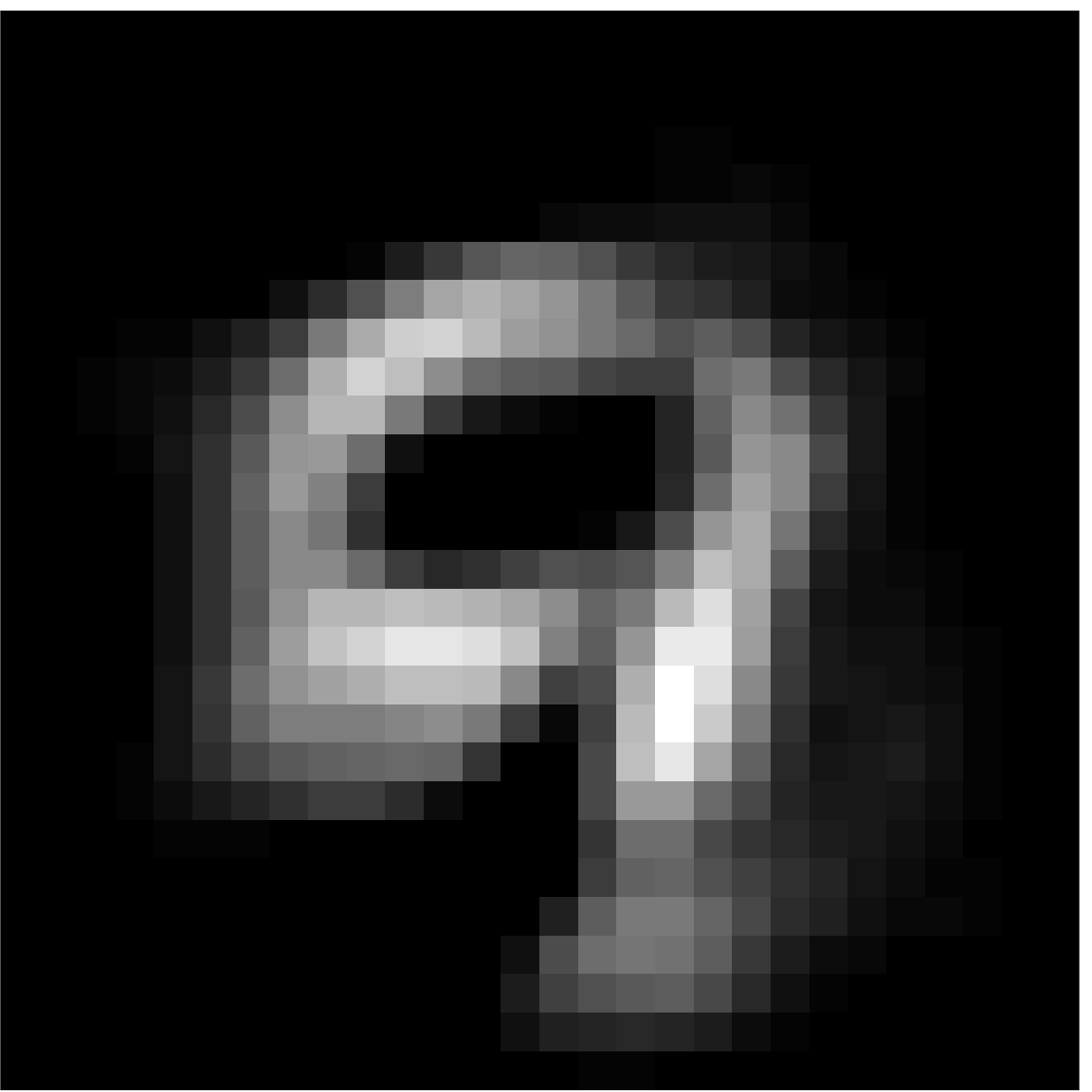}\\
 \caption{The first row shows the Reconstructed Digits 2, 3, 7 and 9 from the learned centroids. The second row shows the reconstructed digits with positive values.}\label{Figure4}
\end{center}
\end{figure}

Figure \ref{Figure4} shows the reconstructed representatives from the learned centroids for $K = 4$, given by $\mathcal{X}^* = \mathcal{G}^*\times_1 \mathbf U_1\times_2 \cdots \times_{N-1}\mathbf U_{N-1}$ with the learned low dimensional centroids $\mathcal{G}^*$.

\begin{table*}[htb]
\caption{Results for the MNIST handwritten digits with 1000 randomly chosen images (100 from each class).}
\begin{center}
\begin{tabular}{|c|c|c|c|c|c|c|c||c|}
\hline
&\multicolumn{1}{c|}{{Evaluation metric}}%
& \multicolumn{1}{c|}{MM}%
& \multicolumn{1}{c|}{ALS}%
& \multicolumn{1}{c|}{B-NFM}
& \multicolumn{1}{c|}{PGM}
& \multicolumn{1}{c|}{DRCC}
& \multicolumn{1}{c||}{PCA-B}
& \multicolumn{1}{c|}{PCA-Multinomial}\\
 \hline
$K=3$ &AC & \textbf{0.9173} & 0.9133  &  0.8820  & 0.9160  &  0.9013 & 0.8506  &  0.9114\\
 &&(2.800e-4) & (2.6111e-4)& (1.644e-4)& (1.800e-4)& (3.589e-4)& (5.274e-3)& (1.700e-4)\\
&NMI & \textbf{0.7351} & 0.7344  & 0.6806  & 0.7307 & 0.7115  &   0.6231 & 0.7252 \\
 && (2.488e-3) & (1.654e-3)& (6.069e-4)& (1.792e-3)& (1.704e-3)& (1.492e-2)& (9.761e-4)\\ \hline
$K=4$ &AC  & 0.7140 &0.6520  & 0.6780   & \textbf{0.7195}  & 0.6575  & 0.6670  &  0.6800\\
 && (8.518e-4) & (8.611e-3)& (4.138e-4)& (5.325e-4)& (3.237e-3)& (4.576e-3)& (4.906e-4)\\
&NMI & 0.4654 &0.4181  & 0.4446  & \textbf{0.4680} & 0.4071  & 0.4184   & {0.4360}\\
 &&(1.726e-3) &(3.279e-3)& (5.964e-4)& (1.181e-3)& (3.278e-3)& (2.588e-3)& (7.243e-4)\\ \hline
$K=5$ &AC  & 0.6216 &0.7112  & 0.6104  & 0.6764  & 0.6632  & 0.6960   & \textbf{0.7196 }\\
 && (7.731e-3) &(3.292e-4)& (3.839e-3)& (1.533e-3)& (2.792e-4)& (1.370e-3)& (1.334e-3)\\
&NMI & 0.5201 & \textbf{0.5590}  & 0.4992  & 0.5373 & 0.5198  & 0.5242  &  \textbf{0.5547}\\
 &&(2.080e-3) &(1.018e-3)& (2.921e-3)& (9.196e-4)& (2.156e-4)& (1.130e-3)& (5.525e-4)\\  \hline
$K=6$ &AC  &0.5370 &0.5337  & 0.5156  & 0.5466  & 0.4983  & 0.5283   & \textbf{0.5540}\\
 && (4.249e-3) & (3.626e-3)& (3.812e-3)& (4.257e-3)& (1.061e-3)& (5.234e-3)& (6.386e-3)\\
&NMI & 0.4007 & 0.3852  & 0.3980  & 0.3962 & 0.3670  & 0.3850 &  \textbf{0.4083}\\
 &&(9.891e-4) &(6.482e-4)& (1.197e-3)& (1.507e-3)& (1.348e-3)& (3.077e-3)& (1.945e-3)\\ \hline
$K=7$ &AC  & 0.5988&0.5974  & 0.5762   & 0.5940  & \textbf{0.6585}  & 0.6074  &   {0.6231} \\
 &&(8.816e-5) &(3.295e-3)& (1.051e-3)& (3.704e-3)& (1.789e-3)& (3.704e-3)& (5.618e-3)\\
&NMI & 0.5382 &0.5378  & 0.4972  & 0.5342 & \textbf{0.5629}  & 0.5097  &  {0.5401}\\
 && (7.593e-4) & (1.607e-3)& (7.238e-4)& (9.776e-4)& (9.366e-4)& (1.204e-3)& (3.828e-4)\\ \hline
$K=8$ &AC  & 0.6455&0.6162  & 0.5930   & 0.5942  & {0.6270} & 0.5923  &   \textbf{0.6485}  \\
 && (3.217e-3) &(8.523e-4)& (3.880e-3)& (3.267e-3)& (2.312e-3)& (4.726e-3)& (4.496e-3)\\
&NMI &0.5427 & \textbf{0.5496} & 0.4979  & 0.5152 & {0.5033} & 0.4944  &  0.5254 \\
 && (5.694e-4)&(6.968e-4)& (1.329e-3)& (2.243e-3)& (3.711e-3)& (2.286e-3)& (2.367e-3)\\ \hline
$K=9$ &AC  & 0.5211 &0.5567  & 0.5137   & 0.5302  & 0.5240  & 0.5224  &   \textbf{0.5756 } \\
 &&(1.251e-3) &(8.173e-4)& (5.528e-4)& (1.816e-3)& (7.867e-4)& (8.213e-4)& (3.569e-3)\\
&NMI &0.4621 & 0.4716   &   0.4343  & 0.4720 & 0.4493  & {0.4449 } &   \textbf{0.4774} \\
 && (3.209e-4)&(3.473e-4)& (3.890e-4)& (5.279e-4)& (8.811e-4)& (1.130e-3)& (5.103e-4)\\ \hline
$K=10$ &AC &0.4980& 0.4800  & 0.4474   & 0.4958  & 0.4668  & 0.4800  &   \textbf{0.5126 }\\
 && (1.154e-3) &(9.220e-4)& (5.823e-4)& (1.833e-3)& (2.846e-3)& (2.761e-3)& (1.785e-3)\\
&NMI & 0.4470 &0.4505   & 0.4089  & \textbf{0.4516}  & 0.4053  & 0.4142  &  {0.4461}\\
 && (7.543e-4) &(5.267e-4)& (3.474e-4)& (2.051e-3)& (3.402e-3)& (8.642e-4)& (1.146e-3)\\ \hline
\end{tabular}
\label{Table3}
\end{center}
\end{table*}

\subsubsection*{Experiment III}
A test similar to Experiment II is conducted on the PIE dataset to further confirm the main conclusion form Experiment II, that the performance of PCA-Multinomial is better when are a larger number of clusters are sought. We \Gao{do not report} the performance of the Tri-ONTD algorithm as it does not provide meaningful results in our experiments. The PIE dataset contains $68$ clusters, each with $42$ objects. In this experiment, we \Gao{test} the algorithms on the data of $68$, $58$, $48$, $38$, $28$, $18$, and $8$ clusters, respectively. To maintain the data size to be around $600$, we randomly \Gao{pick} $9$ images for the case of cluster $68$, $11$ for $58$, $13$ for $48$, $16$ for $38$, $21$ for $28$, $33$ for $18$, and $42$ for $8$. The procedure is repeated for $5$ runs. In the last case, the data size is $336$ and each run is with $8$ different clusters.

The results are collected in Table \ref{Table4}. PCA-Multinomial performs better than the others, except for the case  $K=8$, where MM takes the lead. Part of this behavior is due to \emph{lack of sufficient data} in the case of cluster $K=8$. However NMI scores are all over $80\%$, except for the case $K=8$. Both DRCC and PCA-B do not show meaningful results on this dataset.
\begin{table*}[htb]
\caption{Results on PIE face dataset with about 600 randomly chosen data.}
\begin{center}
\begin{tabular}{|c|c|c|c|c|c|c|c||c|}
\hline
&\multicolumn{1}{c|}{{Evaluation metric}}
& \multicolumn{1}{c|}{MM}
& \multicolumn{1}{c|}{ALS}
& \multicolumn{1}{c|}{B-NFM}
& \multicolumn{1}{c|}{PGM}
& \multicolumn{1}{c|}{DRCC}
& \multicolumn{1}{c||}{PCA-B}
& \multicolumn{1}{c|}{PCA-Multinomial}\\
 \hline
$K=68$ &AC  & 0.6088   & 0.5633   & 0.6192  & 0.5120  & 0.2745 &   0.4314  & \textbf{0.6533}\\
&& (4.563e-4) & (2.101e-4) & (4.311e-4) & (7.590e-4) & (2.736e-4) & (1.018e-3) & (8.749e-4)\\
&NMI & 0.8066   & 0.7791  & 0.8099  & 0.7540  & 0.5652  &   0.6828  & \textbf{0.8634}\\
&& (2.222e-4) & (4.083e-5) & (9.111e-5) & (9.963e-5) & (3.231e-4) & (3.197e-4) & (2.484e-4)\\
\hline
$K=58$ &AC  & {0.6182}  & 0.5730  & 0.5968  & 0.5482  & 0.2504  &   {0.4125} & \textbf{0.6781}\\
&& (5.719e-4) & (3.549e-3) & (1.884e-4) & (1.362e-3) & (2.215e-4) & (1.318e-3) & (9.944e-4)\\
&NMI &  {0.8027 }  &   0.7790  & 0.7955 & 0.7580  &  0.5140  &   0.6555 &  \textbf{0.8742}\\
&& (2.504e-4) & (6.398e-4) & (2.789e-5) & (2.096e-4) & (2.995e-4) & (8.198e-4) & (1.483e-4)\\
\hline
$K=48$ &AC & 0.6112   & 0.5964   & 0.6147  & 0.5951  & 0.2875  &   {0.4228} & \textbf{0.6990}\\
&& (1.625e-4) & (6.027e-4) & (1.777e-3) & (5.105e-4) & (4.925e-4) & (6.271e-4) & (7.863e-4)\\
&NMI & 0.7899  & 0.7812  & 0.7896 & 0.7775  & 0.5266  &   {0.6491 }& \textbf{0.8480}\\
&& (1.084e-4) & (4.319e-4) & (1.928e-4) & (1.633e-4) & (2.659e-4) & (1.115e-4) & (8.585e-5)\\
\hline
$K=38$ &AC  & {0.5799}  & 0.6197   & 0.6138  &  0.5930  & 0.2621  &   0.3911 &  \textbf{0.7118}\\
&& (2.292e-3) & (9.405e-4) & (9.892e-4) & (3.646e-4) & (3.578e-4) & (1.227e-3) & (1.384e-3)\\
&NMI & 0.7552  & 0.7769  & 0.7717 & 0.7553  & 0.4461  &  0.5949 &  \textbf{0.8516}\\
&& (3.956e-4) & (6.884e-4) & (5.941e-4) & (1.132e-4) & (1.461e-4) & (5.438e-4) & (1.716e-4)\\
\hline
$K=28$ &AC & 0.6224   & 0.6214   & 0.6329  & 0.6408  & 0.2619 &   0.4212 & \textbf{0.6877}\\
&& (2.564e-3) & (1.448e-3) & (4.329e-4) & (8.523e-4) & (2.704e-4) & (4.090e-3) & (4.106e-3)\\
&NMI &  0.7470  & 0.7610  & 0.7586 & 0.7519  & 0.4033 &  0.5669  & \textbf{0.8153}\\
&& (4.750e-4) & (5.707e-4) & (1.975e-4) & (3.880e-4) & (3.592e-4) & (2.179e-3) & (5.873e-4)\\
\hline
$K=18$ &AC  & {0.6367}  & {0.6387}  & 0.5892  & 0.6511  & 0.2962  &   0.4865  & \textbf{0.7329}\\
&& (6.368e-4) & (1.002e-2) & (5.057e-3) & (2.232e-3) & (2.326e-3) & (6.932e-3) & (1.352e-3)\\
&NMI & {0.7290} & 0.7433  & 0.7089 & 0.7377  & 0.3836  &  0.5917 & \textbf{0.8010}\\
&& (5.039e-4) & (2.437e-3) & (8.551e-4) & (4.477e-4) & (1.570e-3) & (3.972e-3) & (7.787e-4)\\
\hline
$K=8$ &AC & \textbf{0.6690}& {0.6256}  & 0.6321  & 0.6285  & 0.3809  &   0.4785  & {0.6523}\\
&& (1.073e-3) & (2.698e-3) & (9.592e-4) & (1.396e-3) & (9.553e-3) & (3.036e-2) & (1.634e-2)\\
&NMI &  0.6512  & {0.6797 }& 0.6382 & 0.6289  & 0.3110  &  0.4403  & \textbf{0.6803 }\\
&& (4.287e-3) & (5.646e-4) & (6.623e-4) & (1.509e-3) & (1.269e-2) & (3.941e-2) & (1.057e-2)\\
\hline
\end{tabular}
\label{Table4}
\end{center}
\end{table*}

\subsubsection*{Experiment IV}
In this experiment, we \Gao{test} the algorithms on both the ORL dataset of $40$ subjects with $400$ images and the YaleB extended dataset of $38$ subjects with $2414$ images.

For the ORL dataset, we randomly \Gao{choose} $10$ subjects for five runs of the algorithm. The results in Table \ref{Table5} show that the PCA-Multinomial algorithm outperforms  all the other algorithms on the ORL dataset in terms of both AC and NMI scores.

For the YaleB extended dataset, we \Gao{use} all the images. Hence, only one test \Gao{is} conducted and the results are summarized in Table \ref{Table5}. It should be noted that all the algorithms \Gao{show} poor performance on the YaleB dataset although the PCA-Multinomial algorithm has relatively better performance. This is due to the large variants among this dataset.
\begin{table*}[htb]
\caption{Results for both ORL and YaleB face datasets.}
\begin{center}
\begin{tabular}{|c|c|c|c|c|c|c|c||c|}
\hline
&\multicolumn{1}{c|}{{Evaluation metric}}%
& \multicolumn{1}{c|}{MM}%
& \multicolumn{1}{c|}{ALS}%
& \multicolumn{1}{c|}{B-NFM}
& \multicolumn{1}{c|}{PGM}
& \multicolumn{1}{c|}{DRCC}
& \multicolumn{1}{c||}{PCA-B}
& \multicolumn{1}{c|}{PCA-Multinomial}\\
 \hline
ORL &AC  & 0.6440 & 0.6820  & 0.6280 & 0.6300 & 0.5580 &  0.6360 &  \textbf{0.7340}\\
&& (3.730e-3) &(9.070e-3) & (8.070e-3) & (3.650e-3) & (4.870e-3) & (5.130e-3) & (2.230e-3)\\
    &NMI     & 0.7308  & 0.7269  & 0.7107 & 0.7236 & 0.6498 &  0.6968 & \textbf{0.7996}\\
&& (1.842e-3) &(5.946e-3) & (2.688e-3) & (1.668e-3) & (8.727e-4) & (5.467e-3) & (7.283e-4)
\\ \hline \hline
YaleB &AC  & 0.2175  & 0.2270  & 0.2104 & 0.2084 & 0.1002 &  0.2117 & \textbf{0.3293}\\
      &NMI     & 0.3534  & 0.3470  & 0.3508 & 0.3477 & 0.1455 &  0.3402 & \textbf{0.4772}
\\ \hline
\end{tabular}
\label{Table5}
\end{center}
\end{table*}

\Gao{
\subsubsection*{Experiment V}
In this experiment, we test the PCA-Multinomial algorithm against the benchmark PCA-B algorithm on the DynTex++ dataset. The other algorithms are not appropriate for this dataset. Tri-ONTD works for matrix data, i.e., it can handle a set of images or $3$-order tensors, whereas the rest work with vectorial data, i.e., can handle only matrices. The core tensor size is fixed to $(30,30,30)$. We conduct tests on the cases of up to $12$ classes. In each test, the total number of training video \emph{subsequences} is determined to $200$ for $2$ classes, $270$ for $3$ classes, $360$ for $4$ classes, $480$ for $6$ classes, $600$ for $8$ classes and $10$ classes, and $660$ for $12$ classes. The classes are randomly chosen. For each case, five runs are conducted by randomly choosing video subsequences in the chosen classes, except for the case of two classes, where the only $200$ subsequences are all used in a single run.

The results are summarized in Table \ref{Table6}. Particularly, the results show the challenging nature of clustering dynamic textures for $K\geq 6$ classes. In all the cases, PCA-Multinomial algorithm consistently performs better than the benchmark PCA-B. This once again confirms the benefits of using the RTR algorithm based on the  Riemannian geometry of the multinomial manifold.}
\begin{table*}[htb]
\caption{Results for the DynTex++ dataset.}
\begin{center}
\begin{tabular}{|c|c|c|c|c|c|c|c|c|}
\hline
\multicolumn{1}{|c|}{{Models}}
& \multicolumn{1}{|c|}{{Evaluation metric}}
& \multicolumn{1}{c|}{2 Classes}
& \multicolumn{1}{c|}{3 Classes}
& \multicolumn{1}{c|}{4 Classes}
& \multicolumn{1}{c|}{6 Classes}
& \multicolumn{1}{c|}{8 Classes}
& \multicolumn{1}{c|}{10 Classes}
& \multicolumn{1}{c|}{12 Classes}\\
 \hline
PCA-B &AC  & 0.5550  & 0.4230  & 0.3727  & 0.2983  & 0.2507  & 0.1937   &  {0.2024}\\
&& - &( 4.069e-3) & ( 1.429e-3) & ( 4.088e-4) & ( 5.244e-4) & ( 8.527e-5) & (3.416e-4)\\
    &NMI     &  0.1034  & 0.1849   & 0.2261  & 0.2079  & 0.2259  & 0.2451   &  {0.2520}\\
&& - &( 8.471e-3) & ( 1.667e-3) & ( 4.817e-4) & ( 2.926e-4) & ( 1.057e-3) & (9.549e-5)
\\ \hline \hline
PCA-Multinomial &AC  &  \textbf{0.7901}  &  \textbf{0.5827}  & \textbf{0.5767}  &\textbf{0.3842}   &\textbf{0.4007}   & \textbf{0.3300}   & \textbf{0.2980}\\
&& - &( 5.522e-4) & ( 6.350e-3) & ( 1.361e-3) & ( 1.325e-3) & ( 1.272e-3) & ( 2.519e-4)  \\
&NMI     &  \textbf{0.4056}  &  \textbf{0.2047} & \textbf{0.3360}  & \textbf{0.3012}  & \textbf{0.3538}  & \textbf{0.3285}   & \textbf{0.3444 }\\
&& - &( 5.924e-4) & ( 6.116e-4) & ( 9.165e-4) & ( 5.356e-4) & ( 4.801e-4) & ( 1.563e-4)
\\ \hline
\end{tabular}
\label{Table6}
\end{center}
\end{table*}

\section{Conclusion}\label{Sec:6}
We \Gao{proposed} a heterogeneous Tucker decomposition model for tensor clustering. The model simultaneously conducts dimensionality reduction and \Gao{optimizes} membership representation. The dimensionality reduction is conducted along the first $(N-1)$-modes while \Gao{optimizing} membership on the last mode of tensorial data. The resulting optimization problem is a nonlinear optimization problem over the special matrix multinomial manifold. To apply the Riemannian trust region algorithm to the problem, we \Gao{explored} the Riemannian geometry of the manifold under the Fisher information metric. Finally, we implemented the clustering algorithms based on the Riemannian optimization. We used a number of real-world datasets to test the proposed algorithms and compared them with existing state-of-the-art nonnegative matrix factorization methods. Numerical results illustrate the good clustering performance of the proposed algorithms particularly in the case of higher class numbers.

Further work can be carried out in several different directions. For example, in the current work we use an alternating procedure to optimize over all the factor matrices $\mathbf U_1$, $\mathbf U_2$, $\ldots$, $\mathbf U_N$. However, this is not the only way to solve the problem. As the constraint $\mathbf U^T_{n}\mathbf U_n=\mathbf I$ for $n=1,2,\dots,N-1$ defines the Stiefel manifold, the overall optimization problem \eqref{30March2014-4} is defined over the product manifold of $(N-1)$ Stiefel manifolds and one pn the multinomial manifold. This can be directly optimized over the entire product of $N$ manifolds, e.g., using Manopt. \Gao{Another issue that needs to be further investigated is how to scale the algorithm for \eqref{31March2014-6} to high-dimensional data.}

\ifCLASSOPTIONcompsoc
\section*{Acknowledgments}
\else
\section*{Acknowledgment}
\fi
This research was supported under Australian Research Council's Discovery Projects funding scheme (project number DP130100364). This work is also partially supported by the National Natural Science Foundation of China (No. 61370119, 61133003 and 61390510) and the  Natural Science Foundation of Beijing (No. 4132013). \changeBM{Bamdev Mishra is supported as an FRS-FNRS research fellow (Belgian Fund for Scientific Research)}.

\ifCLASSOPTIONcaptionsoff
  \newpage
\fi
\bibliographystyle{IEEEtran}

\end{document}